\documentclass[10pt,journal,compsoc]{IEEEtran}

\usepackage{graphicx}
\usepackage{amsmath,amsthm,amssymb}
\usepackage{dblfloatfix}
\usepackage{booktabs}
\usepackage{enumitem}

\newtheorem{proposition}{Proposition}
\newcommand{\acronym}{BiLRP}
\newcommand{\HessProd}{Hessian$\kern 0.08em \times \kern 0.08em $Product}
\newcommand{\GradInput}{Gradient$\kern 0.08em \times \kern 0.08em $Input}

\usepackage{svg}

\definecolor{darkgray}{RGB}{80,80,80}
\newcommand{\figletter}[1]{{\textbf{ \sffamily \small \color{darkgray} #1.}}}
\begin{document}

\title{
Building and Interpreting Deep Similarity Models
}

\author{Oliver Eberle, Jochen B\"uttner, Florian Kr\"autli, Klaus-Robert M\"uller, Matteo Valleriani, Gr\'egoire Montavon
\IEEEcompsocitemizethanks{

\IEEEcompsocthanksitem O. Eberle is with the Berlin Institute of Technology (TU Berlin), 10587 Berlin, Germany.

\IEEEcompsocthanksitem J. B\"uttner is with the Max Planck Institute for the History of Science, 14159 Berlin, Germany.

\IEEEcompsocthanksitem F. Kr\"autli is with the Max Planck Institute for the History of Science, 14159 Berlin, Germany.

\IEEEcompsocthanksitem K.-R. M\"uller is with the Berlin Institute of Technology (TU Berlin), 10587 Berlin, Germany; the Department of Brain and Cognitive Engineering, Korea University, Seoul 136-713, Korea; and the Max Planck Institut f{\"u}r Informatik, 66123 Saarbr{\"u}cken, Germany. E-mail: klaus-robert.mueller@tu-berlin.de.

\IEEEcompsocthanksitem M. Valleriani is with the Max Planck Institute for the History of Science, 14159 Berlin, Germany.

\IEEEcompsocthanksitem G. Montavon is with the Berlin Institute of Technology (TU Berlin), 10587 Berlin, Germany. E-mail: gregoire.montavon@tu-berlin.de.

}
\thanks{(Corresponding Authors: Gr\'egoire Montavon, Klaus-Robert M\"uller)}}

\IEEEtitleabstractindextext{%
\begin{abstract}
Many learning algorithms such as kernel machines, nearest neighbors, clustering, or anomaly detection, are based on the concept of `distance' or `similarity'. Before similarities are used for training an actual machine learning model, we would like to verify that they are bound to meaningful patterns in the data. In this paper, we propose to make similarities interpretable by augmenting them with an {\em explanation} in terms of input features. We develop \acronym{}, a scalable and theoretically founded method to systematically decompose similarity scores on pairs of input features. Our method can be expressed as a composition of LRP explanations, which were shown in previous works to scale to highly nonlinear functions. Through an extensive set of experiments, we demonstrate that \acronym{} robustly explains complex similarity models, e.g.\ built on VGG-16 deep neural network features. Additionally, we apply our method to an open problem in digital humanities: detailed assessment of similarity between historical documents such as astronomical tables. Here again, \acronym{} provides insight and brings verifiability into a highly engineered and problem-specific similarity model.
\end{abstract}

}

\maketitle

\newcommand{\z}{\boldsymbol{z}}
\newcommand{\x}{\boldsymbol{x}}
\newcommand{\ba}{\boldsymbol{a}}
\newcommand{\rba}{\widetilde{\ba}}
\newcommand{\bx}{\boldsymbol{x}}
\newcommand{\rbx}{\widetilde{\bx}}
\newcommand{\w}{\boldsymbol{w}}

\section{Introduction}

Building meaningful similarity models that incorporate prior knowledge about the data and the task is an important area of machine learning and information retrieval \cite{DBLP:journals/bioinformatics/ZienRMSLM00,DBLP:books/daglib/0021593}. Good similarity models are needed to find relevant items in databases \cite{DBLP:conf/webdb/NiermanJ02,DBLP:conf/ismir/PampalkFW05,DBLP:journals/jcisd/WillettBD98}. Similarities (or kernels) are also the starting point of a large number of machine learning models including discriminative learning \cite{bishop06,DBLP:books/lib/ScholkopfS02}, unsupervised learning \cite{macqueen1967, DBLP:journals/csur/JainMF99, DBLP:journals/pami/ShiM00,DBLP:journals/neco/ScholkopfPSSW01}, and data embedding/visualization \cite{DBLP:journals/neco/ScholkopfSM98,DBLP:conf/nips/MikolovSCCD13,DBLP:journals/ml/MaatenH12}.

An important practical question is how to select the similarity model appropriately. Assembling a labeled dataset of similarities for validation can be difficult: The labeler would need to inspect meticulously multiple pairs of data points and come up with exact real-valued similarity scores. As an alternative, selecting a similarity model based on performance on some proxy task can be convenient (e.g.\ \cite{DBLP:conf/icml/BachLJ04,DBLP:journals/jmlr/SonnenburgRSS06,DBLP:journals/jmlr/WeinbergerS09,DBLP:journals/jmlr/BergstraB12}). In both cases, however, the selection procedure is exposed to a potential lack of representativity of the training data (cf.\ the `Clever Hans' effect \cite{lapuschkin-ncomm19}).---In this paper, we aim for a more direct way to assess similarity models, and make use of explainable ML for that purpose.

\smallskip

Explainable ML \cite{DBLP:series/lncs/11700,DBLP:journals/cacm/Lipton18,DBLP:journals/dsp/MontavonSM18} is a subfield of machine learning that focuses on making predictions interpretable to the human. By highlighting the input features (e.g.\ pixels or words) that are used for predicting, explainable ML allows to gain systematic understanding into the model decision structure. Numerous approaches have been proposed in the context of ML classifiers \cite{DBLP:journals/jmlr/BaehrensSHKHM10,lrp,DBLP:conf/kdd/Ribeiro0G16,DBLP:conf/iccv/SelvarajuCDVPB17}.

\smallskip

\begin{figure}[h]
\centering
\includegraphics[width=.97\linewidth]{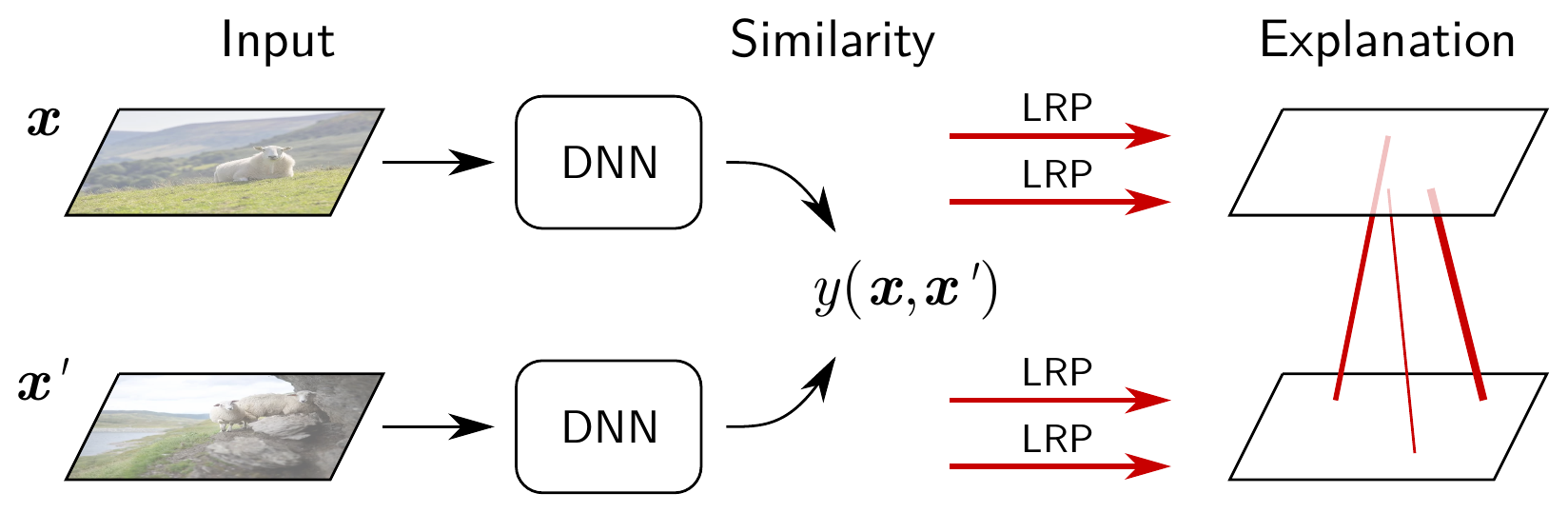}
\caption{Proposed \acronym{} method for explaining similarity. Produced explanations are in terms of pairs of input features.}
\label{fig:intro}
\end{figure}

In this paper, we bring explainable ML to similarity. We contribute a new method that systematically explains similarity models of the type:
$$
y(\x,\x') = \big\langle \phi_L \circ \dots \circ \phi_1(\x)\,,\,\phi_L \circ \dots \circ \phi_1(\x') \big\rangle,
$$
e.g.\ dot products built on some hidden layer of a deep neural network. Our method is based on the insight that similarity models can be naturally decomposed on {\em pairs} of input features. Furthermore, this decomposition can be computed as a combination of multiple LRP explanations \cite{lrp} (and potentially other successful explanation techniques). As a result, it inherits qualities such as broad applicability and scaling to highly nonlinear models. Our method, which we call `\acronym{}', is depicted at a high level in Fig.\ \ref{fig:intro}.

Conceptually, \acronym{} performs a {\em second}-order `deep Taylor decomposition' \cite{DBLP:journals/pr/MontavonLBSM17} of the similarity score, which lets us retrace, layer after layer, features that have jointly contributed to the similarity. Our method reduces for specific choices of parameters to a `\HessProd{}' baseline. With appropriate choices of parameters \acronym{} significantly improves over this baseline and produces explanations that robustly extend to complex deep neural network models.

We showcase \acronym{} on similarity models built at various layers of the well-established VGG-16 image classification network \cite{DBLP:journals/corr/SimonyanZ14a}. Our explanation method brings useful insights into the strengths and limitations of each similarity model. We then move to an open problem in the digital humanities, where similarity between scanned astronomical tables needs to be assessed \cite{mva19}. We build a highly engineered similarity model that is specialized for this task. Again \acronym{} proves useful by being able to inspect the similarity model and validate it from limited data.

Altogether, the method we propose brings transparency into a key ingredient of machine learning: similarity. Our contribution paves the way for designing and validating similarity-based ML models in an efficient, fully informed, and human-interpretable manner.

\subsection{Related Work}

Methods such as LLE \cite{Roweis2000}, diffusion maps \cite{Coifman2006}, or t-SNE \cite{DBLP:journals/ml/MaatenH12} give insight into the similarity structure of large datasets by embedding data points in a low-dimensional subspace where relevant similarities are preserved. While these methods provide useful visualization, their purpose is more to find {\em global} coordinates to comprehend a whole dataset, than to explain why two {\em individual} data points are predicted to be similar.

The question of explaining individual predictions has been extensively studied in the context of ML classifiers. Methods based on occlusions \cite{DBLP:conf/eccv/ZeilerF14,DBLP:conf/iclr/ZintgrafCAW17}, surrogate functions \cite{DBLP:conf/kdd/Ribeiro0G16,DBLP:conf/nips/LundbergL17}, gradients \cite{DBLP:journals/jmlr/BaehrensSHKHM10, DBLP:journals/corr/SimonyanVZ13,DBLP:journals/corr/SmilkovTKVW17, DBLP:conf/icml/SundararajanTY17}, or reverse propagation \cite{lrp,DBLP:conf/eccv/ZeilerF14}, have been proposed, and are capable of highlighting the most relevant features. Some approaches have been extended to unsupervised models, e.g.\ anomaly detection \cite{Kauffmann20,DBLP:conf/icdm/MicenkovaNDA13} and clustering \cite{Kauffmann19}. Our work goes further along this direction and explains {\em similarity} by identifying relevant {\em pairs} of input features.

Several methods for joint features explanations have been proposed. Some of them extract feature interactions globally \cite{DBLP:conf/iclr/TsangC018,kaski-pairwise}. Other methods produce individual explanations for simple pairwise matching models \cite{leupold2017second}, or models with explicit multivariate structures \cite{DBLP:conf/kdd/CaruanaLGKSE15}. Another method extracts joint feature explanations in nonlinear models by estimating the integral of the Hessian \cite{DBLP:journals/corr/abs-2002-04138}. In comparison, our \acronym{} method leverages the layered structure of the model to robustly explain complex similarities, e.g.\ built on deep neural networks.

A number of works improve similarity models by leveraging prior knowledge or ground truth labels. Proposed approaches include structured kernels \cite{Watkins99dynamicalignment,DBLP:journals/bioinformatics/ZienRMSLM00,DBLP:journals/neco/TsudaKRSM02,DBLP:journals/sigkdd/Gartner03}, or siamese/triplet networks \cite{DBLP:conf/nips/BromleyGLSS93,DBLP:conf/cvpr/ChopraHL05,DBLP:conf/cvpr/WangSLRWPCW14,DBLP:conf/simbad/HofferA15,DBLP:conf/eccv/SeguinSdK16}. Beyond similarity, applications such as collaborative filtering \cite{DBLP:conf/www/HeLZNHC17}, transformation modeling \cite{DBLP:journals/neco/MemisevicH10}, and information retrieval \cite{Tzompanaki2012}, also rely on building high-quality matching models between pairs of data.---Our work has an orthogonal objective: It assumes an already trained well-performing similarity model, and makes it explainable to enhance its verifiability and to extract novel insights from it.

\section{Towards Explaining Similarity}
\label{section:towards}

In this section, we present basic approaches to explain similarity models in terms of input features. We first discuss the case of a simple linear model, and then extend the concept to more general nonlinear cases.

\subsection{From Linear to Nonlinear Models}

Let us begin with a simple scenario where $\x,\x' \in \mathbb{R}^d$ and the similarity score is given by some dot product $y(\x,\x') = \langle W \x, W \x' \rangle$, with $W$ a projection matrix of size $h \times d$. The similarity score can be easily decomposed on input features by rewriting the dot product as:
\begin{align}
\textstyle y(\x,\x') =& \textstyle \sum_{ii'} \langle W_{:,i}, W_{:,i'} \rangle \cdot x_i x'_{i'}.
\label{eq:linear}
\end{align}
We observe from Eq.\ \eqref{eq:linear} that the similarity is decomposable on {\em pairs} of features $(i,i')$ of the two examples. In other words, input features interact to produce a high/low similarity score.

In practice, more accurate models of similarity can be obtained by relaxing the linearity constraint. Consider some similarity model $y(\x,\x') = \langle \phi(\x), \phi(\x') \rangle$ built on some abstract feature map $\phi \colon \mathbb{R}^d \to \mathbb{R}^h$ which we assume to be differentiable. A simple and general way of attributing the similarity score to the input features is to compute a Taylor expansion \cite{lrp} at some reference point $(\rbx,\rbx')$:
\begin{align*}
y(\x,\x')
&= y(\rbx,\rbx')\\
&\textstyle \quad + \sum_i \, [\nabla y(\rbx,\rbx')]_{i} \, (x_i - \widetilde{x}_i)\\[1mm]
&\textstyle \quad\quad +  \sum_{i'} \, [\nabla y(\rbx,\rbx')]_{i'} \, (x'_{i'} - \widetilde{x}'_{i'})\\[1mm]
&\textstyle \quad\quad\quad + \sum_{ii'} \, [\nabla^2 y(\rbx,\rbx')]_{ii'} \, (x_i - \widetilde{x}_i) \,(x'_{i'} - \widetilde{x}'_{i'})\\
&\textstyle \quad\quad\quad\quad + \dots
\end{align*}
The explanation is then obtained by identifying the multiple terms of the expansion. Here again, like for the linear case, some of these terms can be attributed to pairs of features $(i,i')$. For general nonlinear models, it is difficult to systematically find reference points $(\rbx,\rbx')$ at which a Taylor expansion represents well the similarity score. To address this, we will need to apply some restrictions to the analyzed model.

\subsection{The `\HessProd{}' Baseline}

Consider the family of similarity models that can be represented as dot products on positively homogeneous feature maps, i.e.\
\begin{align*}
y(\x,\x') &= \langle \phi(\x) , \phi(\x') \rangle,\\
\phi &\colon \mathbb{R}^d \to \mathbb{R}^h \quad \text{with} \quad \forall_{\x}\forall_{t>0} : \phi(t\x) = t\phi(\x).
\end{align*}
The class of functions $\phi$ is broad enough to include (with minor restrictions) interesting models such as the mapping on some layer of a deep rectifier network \cite{DBLP:journals/jmlr/GlorotBB11,DBLP:journals/corr/SimonyanZ14a,DBLP:conf/cvpr/HeZRS16}.

\smallskip

We perform a Taylor expansion of the similarity function at the reference point $(\widetilde{\x},\widetilde{\x}') = (\varepsilon\kern 0.04em \x,\varepsilon\kern 0.04em \x')$ with $\varepsilon$ almost zero. Zero- and first-order terms of the expansion vanish, leaving us with a decomposition on the interaction terms:
\begin{align}
\textstyle y(\x,\x') &= \textstyle \sum_{ii'} [\nabla^2 y(\x,\x')]_{ii'} \, x_i x'_{i'}
\label{eq:hessprod}
\end{align}
(cf.\ Appendix A of the Supplement). Inspection of these interaction terms reveals that a pair of features $(i,i')$ is found to be relevant if:

\begin{enumerate}[label=(\roman*)]

\item the features are jointly expressed in the data, and

\item the similarity model jointly reacts to these features.

\end{enumerate}

\noindent We call this method `\HessProd{}' (HP) and use it as a baseline in Section \ref{section:baselines}. This baseline can also be seen as a reduction of `Integrated Hessians' \cite{DBLP:journals/corr/abs-2002-04138} for the considered family of similarity models.

\smallskip

HP is closely connected to a common baseline method for explaining ML classifiers: \GradInput{} \cite{DBLP:journals/corr/ShrikumarGSK16,DBLP:conf/iclr/AnconaCO018,axioms}. The matrix of joint feature contributions found by HP can be obtained by performing $2\times h$ `\GradInput{}' (GI) computations:
\begin{align*}
\mathrm{HP}(y,\x,\x') &= \sum_{m=1}^h \mathrm{GI}(\phi_m,\x) \otimes \mathrm{GI}(\phi_m,\x')
\end{align*}
(cf.\ Appendix A.3 of the Supplement). This gradient-based formulation makes it easy to implement HP using neural network libraries with automatic differentiation. However, because of this close relation, `\HessProd{}' also inherits some weaknesses of `\GradInput{}', in particular, its high exposure to gradient noise \cite{axioms}. In deep architectures, the gradient is subject to a shattering effect \cite{DBLP:conf/icml/BalduzziFLLMM17} making it increasingly large, high-varying, and uninformative, with every added layer.

\section{Better Explanations with \acronym{}}\label{section:bilrp}

Motivated by the limitations of the simple techniques presented in Section \ref{section:towards}, we introduce our new \acronym{} method for explaining similarities. The method is inspired by the `layer-wise relevance propagation' (LRP) \cite{lrp} method, which was first introduced for explaining deep neural network classifiers. LRP leverages the layered structure of the model to produce robust explanations.

\smallskip

\acronym{} brings the robustness of LRP to the task of explaining dot product similarities. Our method assumes as a starting point a layered similarity model:
$$
y(\x,\x') = \big\langle \phi_L \circ \dots \circ \phi_1(\x)\,,\,\phi_L \circ \dots \circ \phi_1(\x') \big\rangle,
$$
typically, a dot product built on some hidden layer of a deep neural network. Similarly to LRP, the model output is propagated backward in the network, layer after layer, until the input features are reached. \acronym{} operates by sending messages  $R_{jj'\leftarrow kk'}$ from pairs of neurons $(k,k')$ at a given layer to pairs of neurons $(j,j')$ in the layer below.

\subsection{Extracting \acronym{} Propagation Rules}
\label{section:bilrp-derivation}

\begin{figure}[t]
\centering
\includegraphics[width=0.98\linewidth]{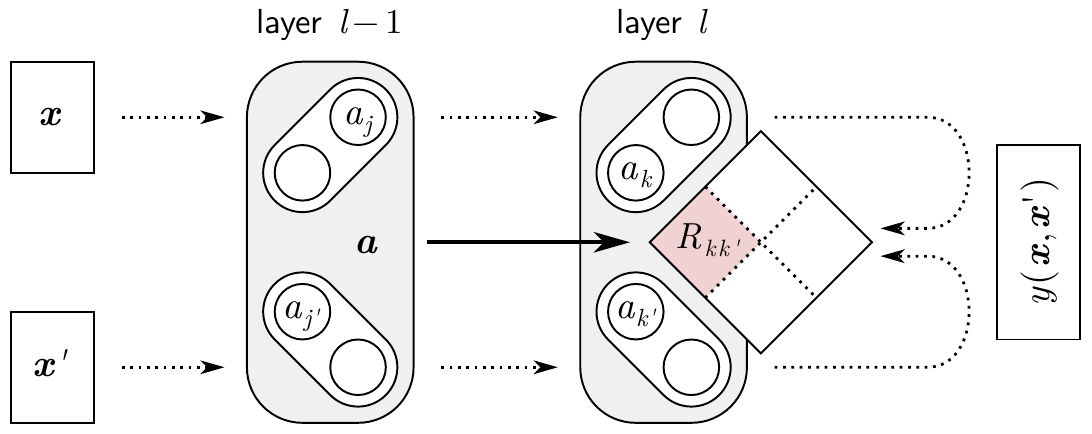}
\caption{Diagram of the map used by DTD to derive \acronym{} propagation rules. The map connects activations at some layer to relevance in the layer above.}
\label{fig:map}
\end{figure}

To build meaningful propagation rules, we make use of the `deep Taylor decomposition' (DTD) \cite{DBLP:journals/pr/MontavonLBSM17} framework. DTD expresses the relevance $R_{kk'}$ available for redistribution as a function of activations $\ba$ in the layer below. The relation between these two quantities is depicted in Fig.\ \ref{fig:map}.
Specifically, DTD seeks to perform a Taylor expansion of the function $R_{kk'}(\ba)$ at some reference point $\rba$:
\begin{align*}
R_{kk'}(\ba)
&= \textstyle R_{kk'}(\rba)\\
& \quad + \textstyle \sum_j [\nabla R_{kk'}(\rba)]_j \cdot (a_j - \widetilde{a}_j) \\
&\quad\quad + \textstyle \sum_{j'} [\nabla R_{kk'}(\rba)]_{j'} \cdot (a_{j'} - \widetilde{a}_{j'})\\
&\quad\quad\quad+ \textstyle \sum_{jj'} [\nabla^2 R_{kk'}(\rba)]_{jj'} \cdot (a_j - \widetilde{a}_j) \,
(a_{j'} - \widetilde{a}_{j'})\\
&\quad\quad\quad\quad+ \dots
\end{align*}
so that messages $R_{jj'\leftarrow kk'}$ can be identified. In practice, the function $R_{kk'}(\ba)$ is difficult to analyze, because it subsumes a potentially large number of forward and backward computations. DTD introduces the concept of a `relevance model' $\widehat{R}_{kk'}(\ba)$ which locally approximates the true relevance score, but only depends on corresponding activations \cite{DBLP:journals/pr/MontavonLBSM17}. For linear/ReLU layers \cite{DBLP:journals/jmlr/GlorotBB11}, we define the relevance model:
\begin{align*}
\widehat{R}_{kk'}(\ba) &=\textstyle
\underbrace{ \textstyle
\big(\sum_{j} a_j w_{jk} \big)^+
}_{a_k}
\, 
\underbrace{ \textstyle
\big(\sum_{j'} a_{j'} w_{j'k'} \big)^+
}_{a_{k'}}
\,
c_{kk'}
\end{align*}
\vskip -1mm
\noindent with $c_{kk'}$ a constant set in a way that $\widehat{R}_{kk'}(\ba) = R_{kk'}$. This relevance model is justified later in Proposition \ref{prop:model}. We now have an easily analyzable model, more specifically, a model that is bilinear on the joint activated domain and zero elsewhere. We search for a root point $\widetilde{\ba}$ at the intersection between the two ReLU hinges and the plane $\{\rba(t,t') \mid t,t' \in \mathbb{R}\}$ where:
\begin{align*}
[\,\rba(t,t')\,]_{j} &= a_{j} - t a_{j} \cdot (1 + \gamma \cdot 1_{w_{jk} > 0}),\\
[\,\rba(t,t')\,]_{j'} &= a_{j'} - t' a_{j'} \cdot (1 + \gamma \cdot 1_{w_{j'k'} > 0})
\end{align*}
with $\gamma \geq 0$ a hyperparameter. This search strategy can be understood as starting with the activations $\ba$, and jointly decreasing them (especially the ones with positive contributions) until $\widehat{R}_{kk'}(\rba)$ becomes zero. Zero- and first-order terms of the Taylor expansion vanish, leaving us with the interaction terms $R_{jj' \leftarrow kk'}$. The total relevance received by $(j,j')$ from neurons in the layer above is given by:
\begin{align}
R_{jj'}
&= \sum_{kk'}\frac{a_j a_{j'} \rho(w_{jk}) \rho(w_{j'k'})}{\sum_{jj'} a_j a_{j'} \rho(w_{jk}) \rho(w_{j'k'})}
 R_{kk'}
 \label{eq:lrpgamma2}
\end{align}
with $\rho(w_{jk}) = w_{jk} + \gamma w_{jk}^+$. A derivation is given in Appendix B.1 of the Supplement. This propagation rule can be seen as a second-order variant of the LRP-$\gamma$ rule \cite{lrpoverview} used for explaining DNN classifiers. It has the following interpretation: A pair of neurons $(j,j')$ is assigned relevance if the following three conditions are met:
\smallskip
\begin{enumerate}[label=(\roman*)]
\item it jointly activates,
\item some pairs of neurons in the layer above jointly react,
\item these reacting pairs are themselves relevant.
\end{enumerate}
\smallskip
In addition to linear/ReLU layers, we would like \acronym{} to handle other common layers such as max-pooling and min-pooling. These two layer types can be seen as special cases of the broader class of {\em positively homogeneous} layers (i.e.\ satisfying $\forall_{\ba} \forall_{t>0}:~a_k(t\kern 0.04em \ba) = t\kern 0.04em a_k(\ba)$). For these layers, the following propagation rule can be derived from DTD:
\begin{align}
R_{jj'} = \sum_{kk'} \frac{a_j a_{j'} [\nabla^2 a_k a_{k'}]_{jj'}}{\sum_{jj'} a_j a_{j'} [\nabla^2 a_k a_{k'}]_{jj'}} R_{kk'}
\label{eq:lrpother2}
\end{align}
(cf.\ Appendix B.2 of the Supplement). This propagation rule has a similar interpretation to the one above, in particular, it also requires for $(j,j')$ to be relevant that the corresponding neurons activate, that some neurons $(k,k')$ in the layer above jointly react, and that the latter neurons are themselves relevant.

\subsection{\acronym{} as a Composition of LRP Computations}
\label{section:bilrp-composition}

A limitation of a plain application of the propagation rules of Section \ref{section:bilrp-derivation} is that we need to handle at each layer a data structure $(R_{kk'})_{kk'}$ which grows quadratically with the number of neurons. Consequently, for large neural networks, a direct computation of these propagation rules is unfeasible. However, it can be shown that relevance scores at each layer can also written in the factored form:
\begin{align*}
R_{kk'} &= \textstyle \sum_{m=1}^h R_{km} R_{k'm}\\
R_{jj'} &= \textstyle \sum_{m=1}^h R_{jm} R_{j'm}
\end{align*}
where $h$ is the dimension of the top-layer feature map, and where the factors can be computed iteratively as:
\begin{align}
R_{jm} &= \sum_k \frac{a_j \rho(w_{jk})}{\sum_j a_j \rho(w_{jk})} R_{km}
\label{eq:lrpgamma}
\end{align}
for linear/ReLU layers, and 
\begin{align}
R_{jm} &= \sum_k \frac{a_j [\nabla a_k]_j}{\sum_j a_j [\nabla a_k]_j} R_{km}
\label{eq:lrpother}
\end{align}
for positively homogeneous layers. The relevance scores that result from applying these factored computations are strictly equivalent to those one would get if using the original propagation rules of Section \ref{section:bilrp-derivation}. A proof is given in Appendix C of the Supplement.

\smallskip

Furthermore, in comparison to the $(\#\,\text{neurons})^2$ computations required at each layer by the original propagation rules, the factored formulation only requires $(\#\,\text{neurons} \times 2h)$ computations. The factored form is therefore especially advantageous when $h$ is low. In the experiments of Section \ref{section:vgg}, we will improve the explanation runtime of our similarity models by adding an extra layer projecting output activations to a smaller number of dimensions.

\smallskip

Lastly, we observe that Equations \eqref{eq:lrpgamma} and \eqref{eq:lrpother} correspond to common rules used by standard LRP. The first one is equivalent to the LRP-$\gamma$ rule \cite{lrpoverview} used in convolution/ReLU layers of DNN classifiers. The second one corresponds to the way LRP commonly handles pooling layers \cite{lrp}. These propagation rules apply independently on each branch and factor of the similarity model. This implies that \acronym{} can be implemented as a combination of multiple LRP procedures that are then recombined once the input layer has been reached:
\begin{align*}
\text{BiLRP}(y,\x,\x') &= \sum_{m=1}^h \text{LRP}([\phi_{L} \circ \dots \circ \phi_1]_m,\x)\\[-2mm]
&\qquad \qquad \quad \otimes \text{LRP}([\phi_{L} \circ \dots \circ \phi_1]_m,\x')
\end{align*}

\begin{figure}[h]
\centering
\includegraphics[width=.98\linewidth]{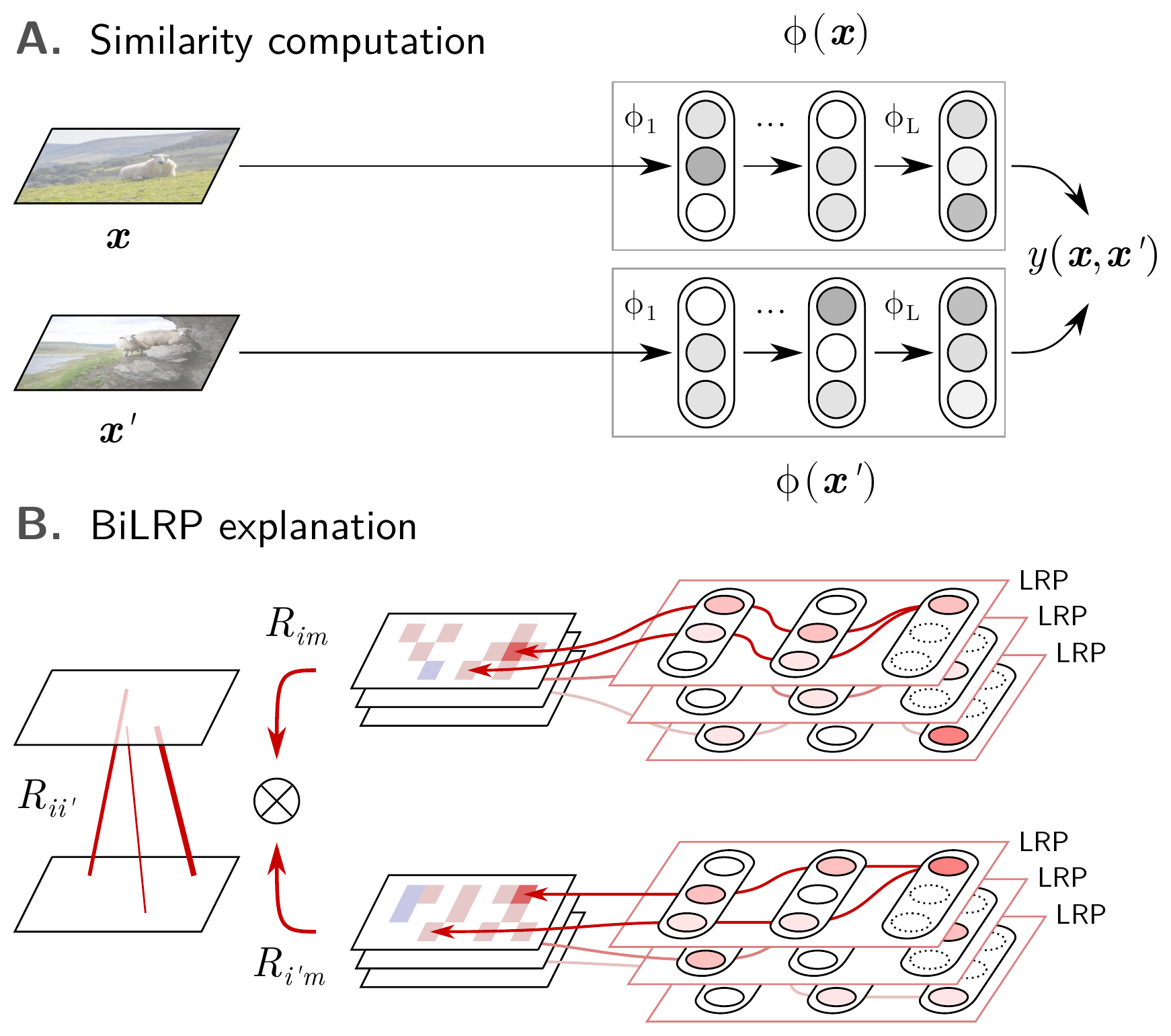}
\caption{Illustration of our approach to compute \acronym{} explanations: \figletter{A} Input examples are mapped by the neural network up to the layer at which the similarity model is built. \figletter{B} LRP is applied to all individual activations in this layer, and the resulting array of explanations is recombined into a single explanation of predicted similarity.}
\label{fig:expflow}
\end{figure}

This modular approach to compute \acronym{} explanations is shown graphically in Fig.\ \ref{fig:expflow}.

With this modular structure, \acronym{} can be easily and efficiently implemented based on existing explanation software. We note that the modular approach described here is not restricted to LRP. Other explanation techniques could in principle be used in the composition. Doing so would however lose the interpretation of the explanation procedure as a deep Taylor decomposition.

\subsection{Theoretical Properties of \acronym{}}

A number of results can be shown about \acronym{}. A first result relates the produced explanation to the predicted similarity. Another result lets us view the \HessProd{} method as a special case of \acronym{}. A last result provides a justification for the relevance models used in Section \ref{section:bilrp-derivation}.

\begin{proposition}
For deep rectifier networks with zero biases, \acronym{} is conservative, i.e.\ $\sum_{ii'}R_{ii'} = y(\x,\x')$.
\end{proposition}

\noindent (See Appendix D.1 of the Supplement for a proof.) Conservation ensures that relevance scores are in proportion to the output of the similarity model.

\begin{proposition}
When $\gamma=0$, explanations produced by \acronym{} reduce to those of \HessProd{}.
\end{proposition}

\noindent (See Appendix D.2 of the Supplement for a proof.) We will find in Section \ref{section:baselines} that choosing non-zero values of $\gamma$ gives better explanations. 

\begin{proposition} The relevance computed by \acronym{} at each layer can be rewritten as $R_{jj'} = a_j a_{j'} c_{jj'}$, where $c_{jj'}$ is locally approximately constant.
\label{prop:model}
\end{proposition}

\noindent (Cf.\ Appendix D.3 of the Supplement.) This property supports the modeling of $c_{jj'}, c_{kk'}, \dots$ as constant, leading to easily analyzable relevance models from which the \acronym{} propagation rules of Section \ref{section:bilrp-derivation} can be derived.

\section{\acronym{} vs.\ Baselines}
\label{section:baselines}

This section tests the ability of the proposed \acronym{} method to produce faithful explanations. In general, ground-truth explanations of ML predictions, especially nonlinear ones, are hard to acquire \cite{DBLP:journals/dsp/MontavonSM18,DBLP:journals/corr/abs-1911-09017}.  Thus, we consider an {\em artificial} scenario consisting of:
\smallskip
\begin{enumerate}[label=(\roman*)]
\item a hardcoded similarity model from which it is easy to extract ground-truth explanations,
\item a neural network trained to reproduce the hardcoded model exactly on the whole input domain.
\end{enumerate}
\smallskip
Because the hardcoded model and the neural network become exact functional copies after training, explanations for their predictions should be the same. Hence, this gives us ground-truth explanations to evaluate \acronym{} against baseline methods.

The hardcoded similarity model takes two random sequences of $6$ digits as input and counts the number of matches between them. The matches between the two sequences form the ground truth explanation. The neural network is constructed and trained as follows: Each digit forming the sequence is represented as vectors in $\mathbb{R}_+^{10}$. To avoid a too simple task, we set these vectors to be correlated. Vectors associated to the digits in the sequence are then concatenated to form an input $\x \in \mathbb{R}_+^{6 \times 10}$. The input goes through two hidden layers of size $100$ and one top layer of size $50$ corresponding to the feature map. We train the network for $10000$ iterations of stochastic gradient descent to minimize the mean square error between predictions and ground-truth similarities, and reach an error of $10^{-3}$, indicating that the neural network solves the problem perfectly.

\begin{figure}[h]
\centering
\includegraphics[width=1.0\linewidth,clip=True,trim=10 0 10 0]{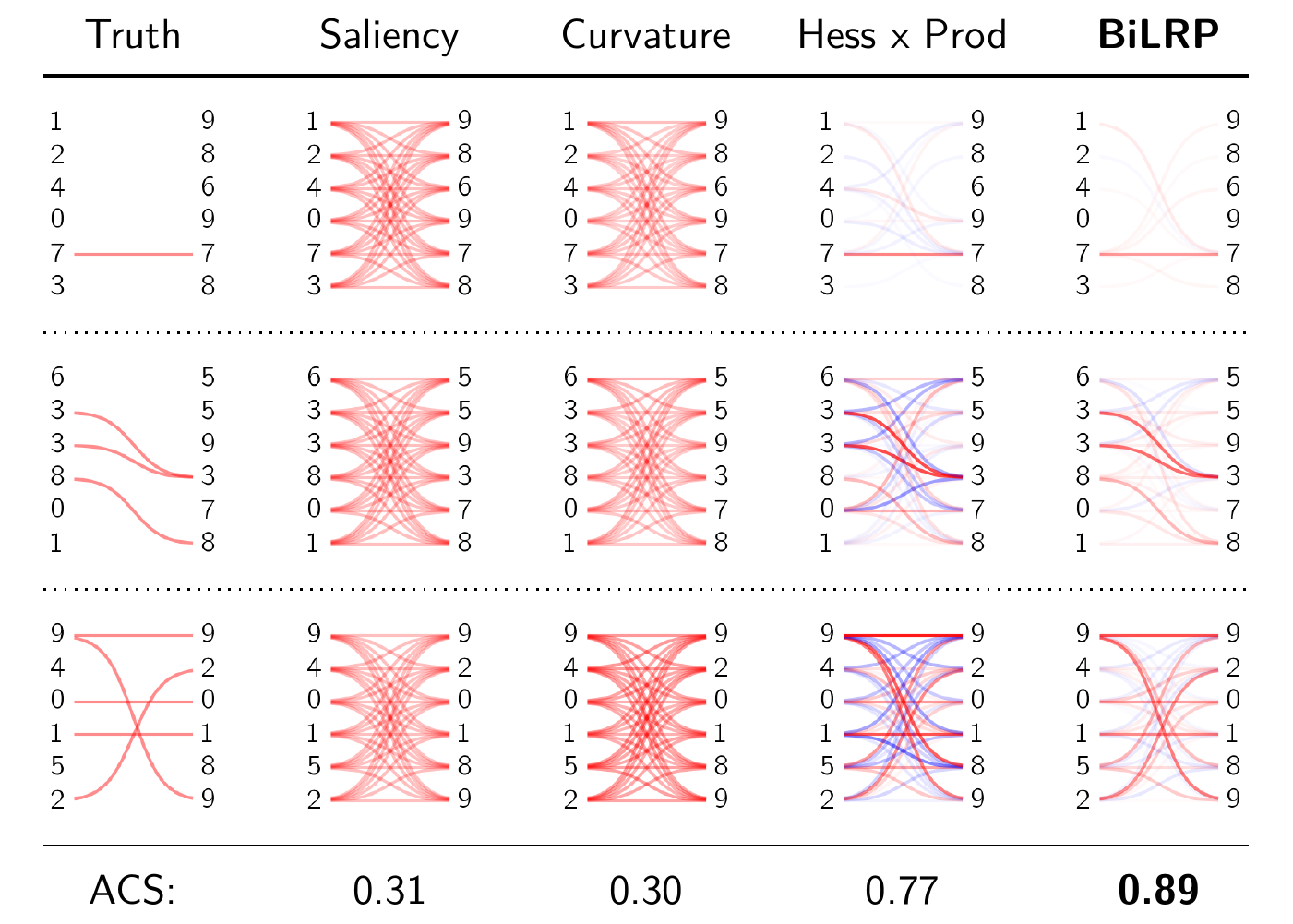}
\caption{Benchmark comparison on a toy example where we have ground-truth explanation of similarity. \acronym{} performs better than all baselines, as measured by the average cosine similarity to the ground truth.}
\label{fig:toy}
\end{figure}

Because there is currently no well-established method for explaining similarity, we consider three simple baselines and use them as a benchmark for evaluating \acronym{}:
\smallskip
\begin{enumerate}[label=--]
\item `Saliency': $R_{ii'} = (x_i x'_{i'})^2$
\vskip 0.5mm
\item `Curvature': $R_{ii'} = ([\nabla^2 y(\x,\x')]_{ii'} )^2$
\vskip 0.5mm
\item `\HessProd{}': $R_{ii'} = x_i x'_{i'}\, [\nabla^2 y(\x,\x')]_{ii'}$
\end{enumerate}
\smallskip
Each explanation method produces a scoring over all pairs of input features, i.e.\ a $(6 \times 10) \times (6 \times 10)$-dimensional explanation. The latter can be pooled over embedding dimensions (cf.\ Appendix E of the Supplement) to form a $6 \times 6$ matrix connecting the digits from the two sequences. Results are shown in Fig.\ \ref{fig:toy}. The closer the produced connectivity pattern to the ground truth, the better the explanation method. High scores are shown in red, low scores in light red or white, and negative scores in blue.

We observe that the `Saliency' baseline does not differentiate between matching and non-matching digits. This is explained by the fact that this baseline is not output-dependent and thus does not know the task. The `Curvature' baseline, although sensitive to the output, does not improve over saliency. The `\HessProd{}' baseline, which can be seen as a special case of \acronym{} with $\gamma=0$, matches the ground truth more accurately but introduces some spurious negative contributions. \acronym{}, through a proper choice of parameter $\gamma$ (here set to $0.09$) considerably reduces these negative contributions.

\begin{figure}[h]
\centering
\includegraphics[width=0.9\linewidth]{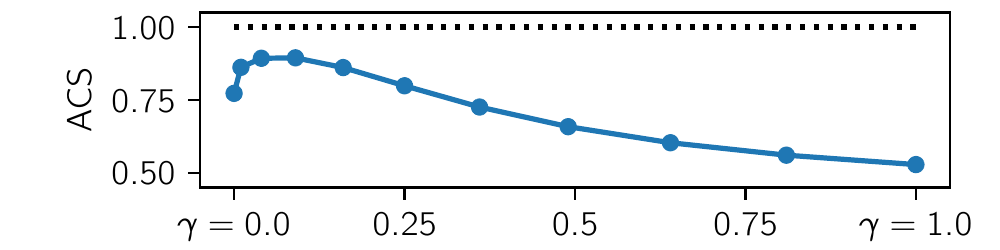}
\caption{Effect of the \acronym{} parameter $\gamma$ on the average cosine similarity between the explanations and the ground truth.}
\label{figure:gamma}
\end{figure}

This visual inspection is validated quantitatively by considering a large number of examples and computing the average cosine similarity (ACS) between the produced explanations and the ground truth. An ACS of 1.0 indicates perfect matching with the ground truth. `Saliency' and 'Curvature' baselines have low ACS. The accuracy is strongly improved by `\HessProd{}' and further improved by \acronym{}. The effect of the parameter $\gamma$ of \acronym{} on the ACS score is shown in Fig.\ \ref{figure:gamma}.

We observe that the best parameter $\gamma$ is small but non-zero. Like for standard LRP, the explanation can be further fine-tuned, e.g.\ by setting the parameter $\gamma$ different at each layer or by considering a broader set of LRP propagation rules \cite{lapuschkin2017faces,lrpoverview}.

\begin{figure*}[t]
\centering
  \includegraphics[width=0.95\linewidth]{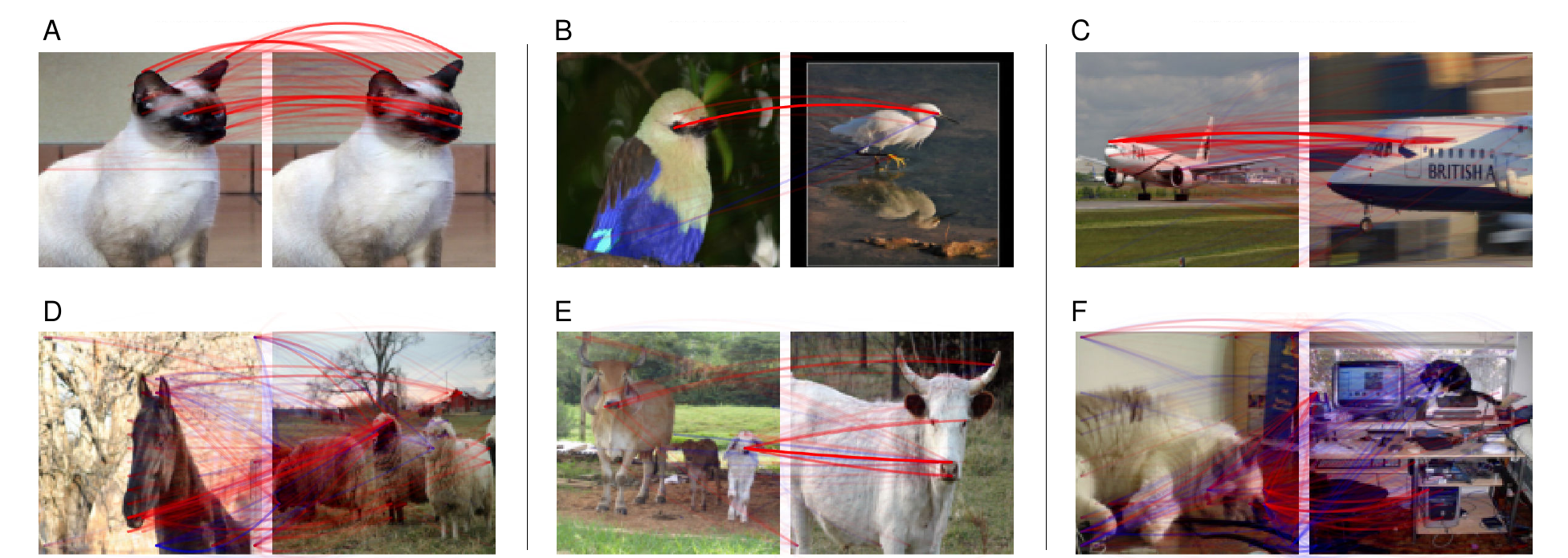}
\caption{Application of \acronym{} to a dot-product similarity model built on VGG-16 features at layer $31$. \acronym{} identifies patterns in the data (e.g.\ ears, eyes) that contribute to the modeled similarity.}
\label{figure:pascal}
\end{figure*}

\section{Interpreting Deep Similarity Models}
\label{section:vgg}

Our next step will be to use \acronym{} to gain insight into practical similarity models built on the well-established \mbox{VGG-16} convolutional neural network \cite{DBLP:journals/corr/SimonyanZ14a}. We take a pretrained version of this network and build the similarity model
\begin{align*}
y(\x,\x') = \big\langle \text{VGG}_{:31}(\x) , \text{VGG}_{:31}(\x') \big\rangle,
\end{align*}
i.e.\ a dot product on the neural network activations at layer $31$. This layer corresponds to the last layer of features before the classifier. The mapping from input to layer $31$ is a sequence of convolution/ReLU layers, and max-pooling layers. It is therefore explainable by \acronym{}. However, the large number of dimensions entering in the dot product computation ($512$ feature maps of size $\frac{w}{32} \times \frac{h}{32}$) where $w$ and $h$ are the dimensions of the input image, makes a direct application of \acronym{} computationally expensive. To reduce the computation time, we append to the last layer a random projection layer that maps activations to a lower-dimensional subspace. In our experiments, we find that projecting to $100$ dimensions provides sufficiently detailed explanations and achieves the desired computational speedup. We set the \acronym{} parameter $\gamma$ to $0.5, 0.25, 0.1, 0.0$ for layers 2--10, 11--17, 18--24, 25--31 respectively. For layer 1, we use the $z^\mathcal{B}$-rule, that specifically handles the pixel-domain \cite{DBLP:journals/pr/MontavonLBSM17}. Finally, we apply a $8 \times 8$ pooling on the output of \acronym{} to reduce the size of the explanations. Details of the rendering procedure are given in Appendix F of the Supplement.

\smallskip

Figure \ref{figure:pascal} shows our \acronym{} explanations on a selection of images pairs taken from the Pascal VOC 2007 dataset \cite{pascal-voc-2007} and resized to $128 \times 128$ pixels. Positive relevance scores are shown in red, negative scores in blue, and score magnitude is represented by opacity. Example A shows two identical images being compared. \acronym{} finds that eyes, nose, and ears are the most relevant features to explain similarity. Example B shows two different images of birds. Here, the eyes are again contributing to the high similarity. In Example C, the front part of the two planes are matched.

Examples D and E show cases where the similarity is not attributed to what the user may expect. In Example D, the horse's muzzle is matched to the head of a sheep. In Example E, while we expect the matching to occur between the two large animals in the image, the true reason for similarity is a small white calf in the right part of the first image. In example F, the scene is cluttered, and does not let appear any meaningful similarity structure, in particular, the two cats are not matched. We also see in this last example that a substantial amount of negative relevance appears, indicating that several joint patterns contradict the similarity score.

Overall, the \acronym{} method gives insight into the strengths and weaknesses of a similarity model, by revealing the features and their relative poses/locations that the model is able or not able to match.

\subsection{How {\em Transferable} is the Similarity Model?}

Deep neural networks, through their multiple layers of representation, provide a natural framework for multitask/transfer learning \cite{DBLP:journals/ml/Caruana97,DBLP:conf/cvpr/OquabBLS14}. DNN-based transfer learning has seen many successful applications \cite{DBLP:conf/kdd/ZhangLZSKYJ15,DBLP:journals/mia/LitjensKBSCGLGS17,DBLP:journals/cacie/GaoM18}. In this section, we consider the problem of transferring a {\em similarity} model to some task of interest. We will use \acronym{} to compare different similarity models, and show how their transferability can be assessed visually from the explanations.

We take the pretrained VGG-16 model and build dot product similarity models at layers $5, 10, 17, 24, 31$ (i.e.\ after each max-pooling layer):
\begin{align*}
y^{(5)}(\x,\x') &= \big\langle \text{VGG}_{:5}(\x) , \text{VGG}_{:5}(\x') \big\rangle,\\[-2mm]
&~~\vdots\\[-2mm]
y^{(31)}(\x,\x') &= \big\langle \text{VGG}_{:31}(\x) , \text{VGG}_{:31}(\x') \big\rangle
\end{align*}
Like in the previous experiment, we add to each feature representation a random projection onto $100$ dimensions in order to make explanations faster to compute. In the following experiments, we consider transfer of similarity to the following three datasets:

\smallskip

\begin{enumerate}[label=--]

\item `Unconstrained Facial Images' (UFI) \cite{DBLP:conf/micai/LencK15},

\item `Labeled Faces in the Wild' (LFW) \cite{LFWTech},

\item `The Sphaera Corpus' \cite{mva19,mva20}.

\end{enumerate}

\smallskip

\noindent The first two datasets are face identification tasks. In identification tasks, a good similarity model is needed in order to reliably extract the closest matches in the training data \cite{DBLP:conf/cvpr/ChopraHL05,DBLP:conf/cvpr/SunWT14}. The third dataset is composed of 358 scanned academic textbooks from the 15th to the 17th century containing texts, illustrations and tables related to astronomical studies. Again, similarity between these entities is important, as it can serve to consolidate historical networks \cite{DBLP:conf/eccv/SeguinSdK16,DBLP:journals/lalc/KrautliV18,Lang18}.

\smallskip

\begin{figure}[h]
\centering
\includegraphics[width=1\linewidth]{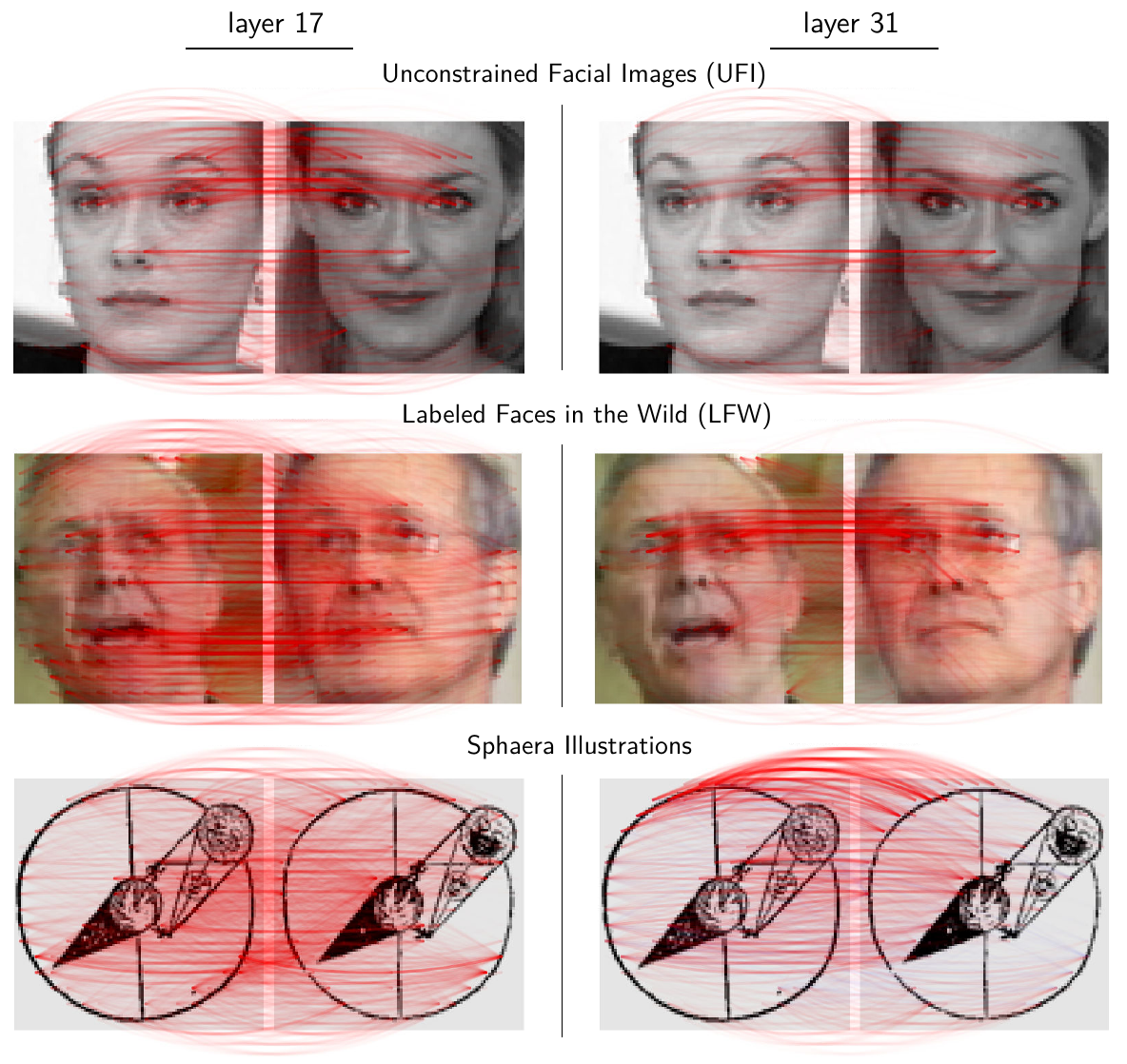}
\caption{Application of \acronym{} to study how VGG-16 similarity transfers to various datasets.}
\label{fig:faces}
\end{figure}

Faces and illustrations are fed to the neural network as images of size $64 \times 64$ pixels and $96 \times 96$ pixels respectively.  We choose for each dataset a pair composed of a test example and the most similar training example. For each pair, we compute the \acronym{} explanations. Results for the similarity model at layer $17$ and $31$ are shown in Fig.\ \ref{fig:faces}.

We observe that the explanation of similarity at layer $31$ is focused on a limited set of features: the eyes or the nose on face images, and a reduced set of lines on the Sphaera illustrations. In comparison, explanations of similarity at layer $17$ cover a broader set of features. These observations suggest that similarity in highest layers, although being potentially capable of resolving very fine variations (e.g.\ for the eyes), might not have kept sufficiently many features in other regions, in order to match images accurately.

To verify this hypothesis, we train a collection of linear SVMs on each dataset where each SVM takes as input activations at a particular layer. On the UFI dataset, we use the original training and test sets. On LFW and Sphaera, data points are assigned randomly with equal probability to the training and test set. The hyperparameter $C$ of the SVM is selected by grid search from the set of values $\{0.001, 0.01, 0.1, 1, 10,100,1000\}$ over $4$ folds on the training set. Test set accuracies for each dataset and layer are shown in Table \ref{table:transfer}.

\begin{table}[h]
\caption{Accuracy of a SVM built on different layers of the VGG-16 network and for different datasets.}
\label{table:transfer}
\centering
\small
\begin{tabular}{lc|ccccc}\toprule
& & \multicolumn{5}{c}{layer}\\[1mm]
dataset & \# classes  & 5 & 10 & 17 & 24 & 31 \\\midrule
UFI & 605  & 0.45 & 0.57 & \bf 0.62  & 0.54 & 0.19 \\
LFW & 61 & 0.78 & 0.86 &\bf 0.92 & 0.89 & 0.75\\
Sphaera & 111 &  0.93 & 0.96 &  \bf 0.98  &  0.97 & 0.96 \\
\bottomrule
\end{tabular}
\end{table}

These results corroborate the hypothesis initially constructed from the \acronym{} explanations: Overspecialization of top layers on the original task leads to a sharp drop of accuracy on the target task. Best accuracies are instead obtained in the intermediate layers.

\subsection{How {\em Invariant} is the Similarity Model?}

To further demonstrate the potential of \acronym{} for characterizing a similarity model, we consider the problem of assessing its invariance properties. Representations that incorporate meaningful invariance are particularly desirable as they enable learning and generalizing from fewer data points \cite{DBLP:journals/pami/BrunaM13,Chmiela2018}.

Invariance can however be difficult to measure in practice: On one hand, the model should respond equally to the input and its transformed version. On the other hand, the response should be selective \cite{Anselmi2016,DBLP:conf/nips/GoodfellowLSLN09}, i.e.\ not the same for every input. In the context of neural networks, a proposed measure of invariance that implements this joint requirement is the local/global firing ratio \cite{DBLP:conf/nips/GoodfellowLSLN09}. In a similar way, we consider an invariance measure for similarity models based on the local/global similarity ratio:
\begin{align}
\textsc{Inv} = 
\frac{\big\langle y(\x,\x') \big\rangle_{\text{local}}}{\big\langle y(\x,\x') \big\rangle_{\text{global}}}
\label{eq:invariance}
\end{align}
The expression $\langle \cdot \rangle_{\text{local}}$ denotes an average over pairs of transformed points (which our model should predict to be similar), and  $\langle \cdot \rangle_{\text{global}}$ denotes an average over all pairs of points.

\smallskip

We study the layer-wise forming of invariance in the VGG-16 network. We use for this the `UCF Sports Action' video dataset \cite{Rodriguez2008ActionMA, ucfsports2014}, where consecutive video frames readily provide a wealth of transformations (translation, rotation, rescaling, etc.) which we would like our model to be invariant to, i.e.\ produce a high similarity score. Videos are cropped to square shape and resized to size $128 \times 128$. We define $\langle \cdot \rangle_{\text{local}}$ to be the average over pairs of nearby frames in the same video ($\Delta t \leq 5$), and $\langle \cdot \rangle_{\text{global}}$ to be the average over all pairs, also from different videos. Invariance scores obtained for similarity models built at various layers are shown in Table \ref{table:invariance}.

\begin{table}[h]
\caption{Invariance measured by Eq.\ \eqref{eq:invariance} at various layers of the VGG-16 network on the UCF Sports Action dataset.}
\label{table:invariance}
\centering
\small
\begin{tabular}{r|ccccc}\toprule
 & \multicolumn{5}{c}{layer}\\[1mm]
  & 5 & 10 & 17 & 24 & 31 \\\midrule
 \textsc{Inv} & 2.30 & 2.31 & 2.43  & 2.87 & \bf 4.00\\
\bottomrule
\end{tabular}
\end{table}

\begin{figure}[h]
\centering
\includegraphics[width=\linewidth]{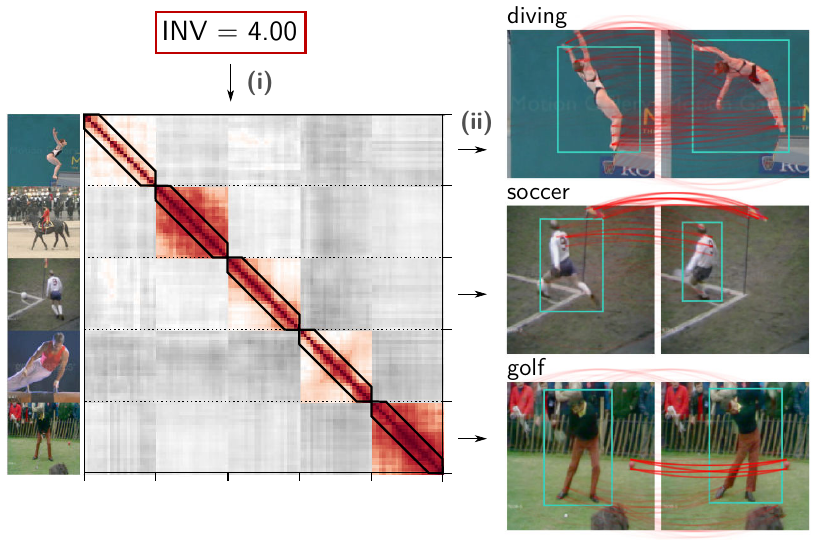}
\caption{Explanation of measured invariance at layer $31$. {\em Left:} Similarity matrix associated to a selection of video clips. The diagonal band outlined in black contains the pairs of examples in $\langle \cdot \rangle_\text{local}$. {\em Right:} \acronym{} explanations for selected pairs from the diagonal band.}
\label{figure:invariance}
\end{figure}

Invariance increases steadily from the lower to the top layers of the neural network and reaches a maximum score at layer $31$. We now take a closer look at the invariance score in this last layer, by applying the following two steps:
\smallskip
\begin{enumerate}[label=(\roman*)]
\item The invariance score is decomposed on the pairs of video frames that directly contribute to it, i.e.\ through the term $\langle \cdot \rangle_\text{local}$ of Eq.\ \eqref{eq:invariance}.
\item \acronym{} is applied to these pairs of contributing video frames in order to produce a finer pixel-wise explanation of invariance.
\end{enumerate}
\smallskip
This two-step analysis is shown in Fig.\ \ref{figure:invariance} for a selection of videos and pairs of video frames.

The first example shows a diver rotating counterclockwise as she leaves the platform. Here, the contribution to invariance is meaningfully attributed to the different parts of the rotating body. The second example shows a soccer player performing a corner kick. Part of the invariance is attributed to the player moving from right to left, however, a sizable amount of it is also attributed in an unexpected manner to the static corner flag behind the soccer player. The last example shows a golf player as he strikes the ball. Again, invariance is unexpectedly attributed to a small red object in the grass. This small object would have likely been overlooked, even after a preliminary inspection of the input images.

The reliance of the invariance measure on unexpected objects in the image (corner flag, small red object) can be viewed as a `Clever Hans' effect \cite{lapuschkin-ncomm19}: the observer assesses how `intelligent' (or invariant) the model is, based on looking at the outcome of a given experiment (the computed invariance score), instead of investigating the decision structure that leads to the high invariance score. This effect may lead to an overestimation of the invariance properties of the model.

\begin{figure}[h]
\centering
  \includegraphics[width=\linewidth]{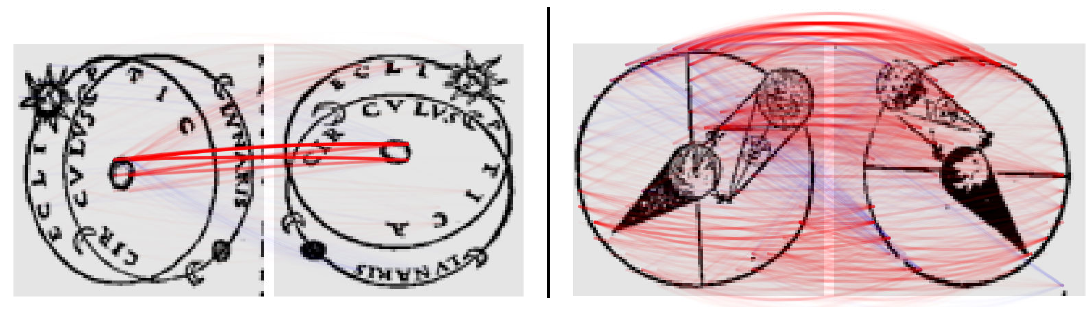}
\caption{Pairs of illustrations from the Sphaera corpus, explained with \acronym{}. The high similarity originates mainly from matching fixed features in the image rather than capturing the rotating elements.}
\label{figure:chans2}
\end{figure}

Similar `Clever Hans' effects can also be observed beyond video data, e.g.\ when applying the similarity model to illustrations in the Sphaera corpus. Figure \ref{figure:chans2} shows two pairs of illustrations whose content is equivalent up to a rotation, and for which our model predicts a high similarity.

Once more, \acronym{} reveals in both cases that the high similarity is not due to matching the rotated patterns, but mainly fixed elements at the center and at the border of the image respectively.

\smallskip

Overall, we have demonstrated that \acronym{} can be useful to identify unsuspected and potentially undesirable reasons for high measured invariance. Practically, applying this method can help to avoid deploying a model with false expectations in real-world applications. Our analysis also suggests that better {\em explanation-based} invariance measures could be designed in the future, potentially in combination with optical flows \cite{DBLP:conf/iccv/DosovitskiyFIHH15}, in order to better distinguish between the matching structures that should and should not contribute to the invariance score.

\section{Engineering Explainable Similarities}
\label{section:engineering}

\begin{figure*}[h]
\centering
  \includegraphics[width=.95\textwidth]{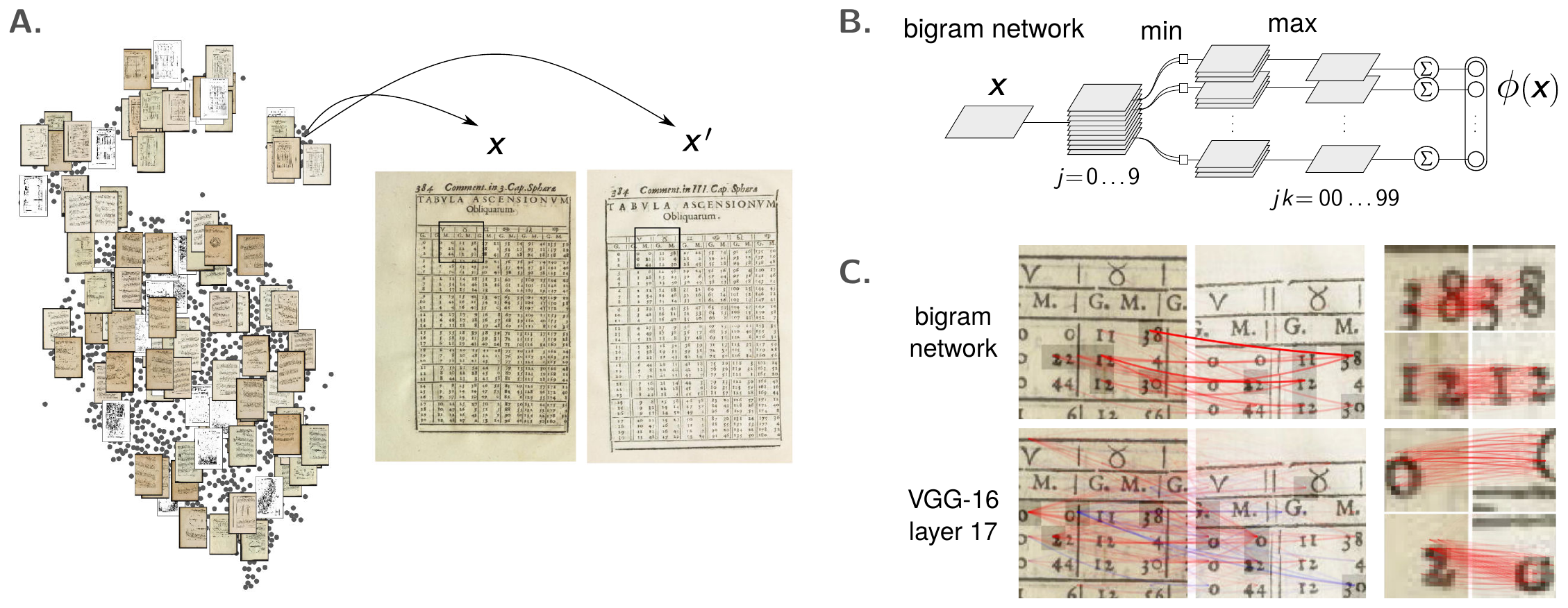}
\caption{\figletter{A} Collection of tables from the Sphaera Corpus \cite{mva19} from which we extract two tables with identical content. \figletter{B}~Proposed `bigram network' supporting the table similarity model. \figletter{C} \acronym{} explanations of predicted similarities between the two input tables.}
\label{figure:sphaera}
\end{figure*}

In this section, we turn to an open and significant problem in the digital humanities: assessing similarity between numeric tables in historical textbooks. We consider scanned numeric tables from the Sphaera Corpus \cite{mva19}. Tables contained in the corpus typically report astronomical measurements or calculations of the positions of celestial objects in the sky. Examples of such tables are given in Fig.\ \ref{figure:sphaera}\,A. Producing an accurate model of similarity between astronomical tables would allow to further consolidate historical networks, which would in turn allow for better inferences.

\smallskip

The similarity prediction task has so far proved challenging: First, it is difficult to acquire ground truth similarity. Getting similarity labels would require a meticulous inspection by a human expert of potentially large tables, and the process would need to be repeated for many of pairs of tables. Also, unlike natural images, faces, or illustrations, which are all well represented by existing pretrained convolutional neural networks, table data usually requires ad-hoc approaches \cite{husson14,DBLP:conf/icdar/SchreiberAWDA17}. In particular, we need to specify which aspects of the tables (e.g.\ numbers, style, or layout) should support the similarity.

\subsection{The `Bigram Network'}
\label{section:bigramnet}

We propose a novel `bigram network' to predict table similarity. Our network can be learned from very few human annotations and is designed to encourage the prediction to be based on relevant numerical features. The network consists of two parts:

\smallskip

The first part is a standard stack of convolution/ReLU layers taking a scanned table $\x$ as input and producing $10$ activation maps $\{\ba_j(\x)\}_{j=1}^{10}$ detecting the digits $0$--$9$. The map $\ba_j(\x)$ is trained to produce small Gaussian blobs at locations where digits of class $j$ are present. The convolutional network is trained on a few hundreds of single digit labels along with their respective image patches. We also incorporate a comparable amount of negative examples (from non-table pages) to correctly handle the absence of digits.

\smallskip

The second part of the network is a hard-coded sequence of layers that extracts task-relevant information from the single-digit activation maps. The first layer in the sequence performs an element-wise `min' operation:
\begin{align*}
\ba_{jk}^{(\tau)}(\x) &= \min\big\{\ba_j(\bx) ,\tau(\ba_k(\bx))\big\}
\end{align*}
The `min' operation be interpreted as a continuous `\textsc{and}' \cite{Kauffmann20}, and tests at each location for the presence of bigrams $jk \in 00$--$99$. The function $\tau$ represents some translation operation, and we apply several of them to produce candidate alignments between the digits forming the bigrams (e.g.\ horizontal shifts of 8, 10, and 12 pixels). We then apply the max-pooling layer:
\begin{align*}
\ba_{jk}(\x) &= \max_\tau \big\{ \ba_{jk}^{(\tau)}(\x) \big\}.
\end{align*}
The `max' operation can be interpreted as a continuous `\textsc{or}', and determines at each location whether a bigram has been found for at least one candidate alignment. Finally, a global sum-pooling layer is applied spatially:
$$
\phi_{jk}(\x) = \big\|\ba_{jk}(\x)\big\|_1
$$
It introduces global translation invariance into the model and produces a $100$-dimensional output vector representing the sum of activations for each bigram. The bigram network is depicted in Fig.\ \ref{figure:sphaera}\,B.

\smallskip

From the output of the bigram network, the similarity score can be obtained by applying the dot product $y(\x,\x') = \langle \phi(\x),\phi(\x') \rangle$. Furthermore, because the bigram network is exclusively composed of convolution/ReLU layers and standard pooling operations, similarities built at the output of this network remain fully explainable by \acronym{}.

\subsection{Validating the `Bigram Network' with \acronym{}}

We come to the final step which is to validate the `bigram network' approach on the task of predicting table similarity. Examples of common validation procedures include precision-recall curves, or the ability to solve a proxy task (e.g.\ table classification) from the predicted similarities. These validation procedures require label information, which is however difficult to obtain for this type of data. Furthermore, when the labeled data is not sufficiently representative, these procedures are potentially affected by the `Clever Hans' effect \cite{lapuschkin-ncomm19}.

\smallskip

In the following, we will show that \acronym{}, through the explanatory feedback it provides, offers a much more data efficient way of performing model validation.
We take a pair of tables $(\x,\x')$, which a preliminary manual inspection has verified to be similar. We then apply \acronym{} to explain:
\smallskip
\begin{enumerate}[label=(\roman*)]
\item the similarity score at the output of our engineered task-specific `bigram network',
\item the similarity score at layer 17 of a generic pretrained VGG-16 network.
\end{enumerate}
\smallskip
For the bigram network, the \acronym{} parameter $\gamma$ is set to $0.5$ at each convolution layer. For the VGG-16 network, we use the same \acronym{} parameters as in Section \ref{section:vgg}. The result of our analysis is shown in Fig.\ \ref{figure:sphaera}\,C.

The bigram network similarity model correctly matches pairs of digits in the two tables. Furthermore, matches are produced between sequences occurring at different locations, thereby verifying the structural translation invariance of the model. Pixel-level explanations further validate the approach by showing that individual digits are matched in a meaningful manner. In contrast, the similarity model built on VGG-16 does not distinguish between the different pairs of digits. Furthermore, part of the similarity score is supported by aspects that are not task-relevant, such as table borders.---Hence, for this particular table similarity task, \acronym{} can clearly establish the superiority of the bigram network over VGG-16.

We stress that this assessment could be readily obtained from a {\em single} pair of tables. If instead we would have applied a validation technique that relies only on similarity scores, significantly more data would have been needed in order to reach the same conclusion with confidence. This sample efficiency of \acronym{} (and by extension any successful explanation technique) for the purpose of model validation is especially important in digital humanities or other scientific domains, where ground-truth labels are typically scarce or expensive to obtain.

\section{Conclusion}

Similarity is a central concept in machine learning that is precursor to a number of supervised and unsupervised machine learning methods. In this paper, we have shown that it can be crucial to get a human-interpretable explanation of the predicted similarity before using it to train a practical machine learning model.

We have contributed a theoretically well-founded method to explain similarity in terms of pairs of input features. Our method called \acronym{} can be expressed as a composition of LRP computations. It therefore inherits its robustness and broad applicability, but extends it to the novel scenario of similarity explanation.

The usefulness of \acronym{} was showcased on the task of understanding similarities as implemented by the VGG-16 neural network, where it could predict transfer learning capabilities and highlight clear cases of `Clever Hans' \cite{lapuschkin-ncomm19} predictions. Furthermore, for a practically relevant problem in the digital humanities, \acronym{} was able to demonstrate with very limited data the superiority of a task-specific similarity model over a generic VGG-16 solution.

Future work will extend the presented techniques from binary towards n-ary similarity structures, especially aiming at incorporating the different levels of reliability of the input features. Furthermore we will use the proposed research tool to gain   insight into large data collections, in particular, grounding historical networks to interpretable domain-specific concepts.

\section*{Acknowledgements}

This work was funded by the German Ministry for Education and Research as BIFOLD -- Berlin Institute for the Foundations of Learning and Data (ref.\ 01IS18025A and ref.\ 01IS18037A), and the German Research Foundation (DFG) as Math+: Berlin Mathematics Research Center  (EXC 2046/1, project-ID: 390685689). This work was partly supported by the Institute for Information \& Communications Technology Planning \& Evaluation (IITP) grant funded by the Korea government (No. 2017-0-00451, No. 2017-0-01779).

\bibliographystyle{IEEEtran}
\bibliography{similarity}

\begin{thebibliography}{10}
\providecommand{\url}[1]{#1}
\csname url@samestyle\endcsname
\providecommand{\newblock}{\relax}
\providecommand{\bibinfo}[2]{#2}
\providecommand{\BIBentrySTDinterwordspacing}{\spaceskip=0pt\relax}
\providecommand{\BIBentryALTinterwordstretchfactor}{4}
\providecommand{\BIBentryALTinterwordspacing}{\spaceskip=\fontdimen2\font plus
\BIBentryALTinterwordstretchfactor\fontdimen3\font minus
  \fontdimen4\font\relax}
\providecommand{\BIBforeignlanguage}[2]{{%
\expandafter\ifx\csname l@#1\endcsname\relax
\typeout{** WARNING: IEEEtran.bst: No hyphenation pattern has been}%
\typeout{** loaded for the language `#1'. Using the pattern for}%
\typeout{** the default language instead.}%
\else
\language=\csname l@#1\endcsname
\fi
#2}}
\providecommand{\BIBdecl}{\relax}
\BIBdecl

\bibitem{DBLP:journals/bioinformatics/ZienRMSLM00}
A.~Zien, G.~R{\"{a}}tsch, S.~Mika, B.~Sch{\"{o}}lkopf, T.~Lengauer, and K.-R.
  M{\"{u}}ller, ``Engineering support vector machine kernels that recognize
  translation initiation sites,'' \emph{Bioinformatics}, vol.~16, no.~9, pp.
  799--807, 2000.

\bibitem{DBLP:books/daglib/0021593}
C.~D. Manning, P.~Raghavan, and H.~Sch{\"{u}}tze, \emph{Introduction to
  information retrieval}.\hskip 1em plus 0.5em minus 0.4em\relax Cambridge
  University Press, 2008.

\bibitem{DBLP:conf/webdb/NiermanJ02}
A.~Nierman and H.~V. Jagadish, ``Evaluating structural similarity in {XML}
  documents,'' in \emph{WebDB}, 2002, pp. 61--66.

\bibitem{DBLP:conf/ismir/PampalkFW05}
E.~Pampalk, A.~Flexer, and G.~Widmer, ``Improvements of audio-based music
  similarity and genre classification,'' in \emph{{ISMIR}}, 2005, pp. 628--633.

\bibitem{DBLP:journals/jcisd/WillettBD98}
P.~Willett, J.~M. Barnard, and G.~M. Downs, ``Chemical similarity searching,''
  \emph{Journal of Chemical Information and Computer Sciences}, vol.~38, no.~6,
  pp. 983--996, 1998.

\bibitem{bishop06}
C.~M. Bishop, \emph{Pattern Recognition and Machine Learning (Information
  Science and Statistics)}.\hskip 1em plus 0.5em minus 0.4em\relax Berlin,
  Heidelberg: Springer-Verlag, 2006.

\bibitem{DBLP:books/lib/ScholkopfS02}
B.~Sch{\"{o}}lkopf and A.~J. Smola, \emph{Learning with Kernels: support vector
  machines, regularization, optimization, and beyond}, ser. Adaptive
  computation and machine learning series.\hskip 1em plus 0.5em minus
  0.4em\relax {MIT} Press, 2002.

\bibitem{macqueen1967}
J.~MacQueen, ``Some methods for classification and analysis of multivariate
  observations,'' in \emph{Proceedings of the Fifth Berkeley Symposium on
  Mathematical Statistics and Probability, Volume 1: Statistics}.\hskip 1em
  plus 0.5em minus 0.4em\relax Berkeley, Calif.: University of California
  Press, 1967, pp. 281--297.

\bibitem{DBLP:journals/csur/JainMF99}
A.~K. Jain, M.~N. Murty, and P.~J. Flynn, ``Data clustering: {A} review,''
  \emph{{ACM} Comput. Surv.}, vol.~31, no.~3, pp. 264--323, 1999.

\bibitem{DBLP:journals/pami/ShiM00}
J.~Shi and J.~Malik, ``Normalized cuts and image segmentation,'' \emph{{IEEE}
  Trans. Pattern Anal. Mach. Intell.}, vol.~22, no.~8, pp. 888--905, 2000.

\bibitem{DBLP:journals/neco/ScholkopfPSSW01}
B.~Sch{\"{o}}lkopf, J.~C. Platt, J.~Shawe-Taylor, A.~J. Smola, and R.~C.
  Williamson, ``Estimating the support of a high-dimensional distribution,''
  \emph{Neural Computation}, vol.~13, no.~7, pp. 1443--1471, 2001.

\bibitem{DBLP:journals/neco/ScholkopfSM98}
B.~Sch{\"{o}}lkopf, A.~J. Smola, and K.-R. M{\"{u}}ller, ``Nonlinear component
  analysis as a kernel eigenvalue problem,'' \emph{Neural Computation},
  vol.~10, no.~5, pp. 1299--1319, 1998.

\bibitem{DBLP:conf/nips/MikolovSCCD13}
T.~Mikolov, I.~Sutskever, K.~Chen, G.~S. Corrado, and J.~Dean, ``Distributed
  representations of words and phrases and their compositionality,'' in
  \emph{{NIPS}}, 2013, pp. 3111--3119.

\bibitem{DBLP:journals/ml/MaatenH12}
L.~van~der Maaten and G.~E. Hinton, ``Visualizing non-metric similarities in
  multiple maps,'' \emph{Machine Learning}, vol.~87, no.~1, pp. 33--55, 2012.

\bibitem{DBLP:conf/icml/BachLJ04}
F.~R. Bach, G.~R.~G. Lanckriet, and M.~I. Jordan, ``Multiple kernel learning,
  conic duality, and the {SMO} algorithm,'' in \emph{{ICML}}, ser. {ACM}
  International Conference Proceeding Series, vol.~69.\hskip 1em plus 0.5em
  minus 0.4em\relax {ACM}, 2004.

\bibitem{DBLP:journals/jmlr/SonnenburgRSS06}
S.~Sonnenburg, G.~R{\"{a}}tsch, C.~Sch{\"{a}}fer, and B.~Sch{\"{o}}lkopf,
  ``Large scale multiple kernel learning,'' \emph{J. Mach. Learn. Res.},
  vol.~7, pp. 1531--1565, 2006.

\bibitem{DBLP:journals/jmlr/WeinbergerS09}
K.~Q. Weinberger and L.~K. Saul, ``Distance metric learning for large margin
  nearest neighbor classification,'' \emph{J. Mach. Learn. Res.}, vol.~10, pp.
  207--244, 2009.

\bibitem{DBLP:journals/jmlr/BergstraB12}
J.~Bergstra and Y.~Bengio, ``Random search for hyper-parameter optimization,''
  \emph{J. Mach. Learn. Res.}, vol.~13, pp. 281--305, 2012.

\bibitem{lapuschkin-ncomm19}
S.~Lapuschkin, S.~W{\"a}ldchen, A.~Binder, G.~Montavon, W.~Samek, and K.-R.
  M{\"u}ller, ``Unmasking {C}lever {H}ans predictors and assessing what
  machines really learn,'' \emph{Nature Communications}, vol.~10, p. 1096,
  2019.

\bibitem{DBLP:series/lncs/11700}
W.~Samek, G.~Montavon, A.~Vedaldi, L.~K. Hansen, and K.-R. M{\"{u}}ller, Eds.,
  \emph{Explainable {AI:} Interpreting, Explaining and Visualizing Deep
  Learning}, ser. Lecture Notes in Computer Science.\hskip 1em plus 0.5em minus
  0.4em\relax Springer, 2019, vol. 11700.

\bibitem{DBLP:journals/cacm/Lipton18}
Z.~C. Lipton, ``The mythos of model interpretability,'' \emph{Commun. {ACM}},
  vol.~61, no.~10, pp. 36--43, 2018.

\bibitem{DBLP:journals/dsp/MontavonSM18}
G.~Montavon, W.~Samek, and K.-R. M{\"{u}}ller, ``Methods for interpreting and
  understanding deep neural networks,'' \emph{Digital Signal Processing},
  vol.~73, pp. 1--15, 2018.

\bibitem{DBLP:journals/jmlr/BaehrensSHKHM10}
D.~Baehrens, T.~Schroeter, S.~Harmeling, M.~Kawanabe, K.~Hansen, and K.-R.
  M{\"{u}}ller, ``How to explain individual classification decisions,''
  \emph{J. Mach. Learn. Res.}, vol.~11, pp. 1803--1831, 2010.

\bibitem{lrp}
S.~Bach, A.~Binder, G.~Montavon, F.~Klauschen, K.-R. M{\"u}ller, and W.~Samek,
  ``On pixel-wise explanations for non-linear classifier decisions by
  layer-wise relevance propagation,'' \emph{PLoS ONE}, vol.~10, no.~7, p.
  e0130140, 07 2015.

\bibitem{DBLP:conf/kdd/Ribeiro0G16}
M.~T. Ribeiro, S.~Singh, and C.~Guestrin, ````{W}hy should {I} trust you?'':
  Explaining the predictions of any classifier,'' in \emph{{KDD}}.\hskip 1em
  plus 0.5em minus 0.4em\relax {ACM}, 2016, pp. 1135--1144.

\bibitem{DBLP:conf/iccv/SelvarajuCDVPB17}
R.~R. Selvaraju, M.~Cogswell, A.~Das, R.~Vedantam, D.~Parikh, and D.~Batra,
  ``Grad-{CAM}: Visual explanations from deep networks via gradient-based
  localization,'' in \emph{{ICCV}}.\hskip 1em plus 0.5em minus 0.4em\relax
  {IEEE} Computer Society, 2017, pp. 618--626.

\bibitem{DBLP:journals/pr/MontavonLBSM17}
G.~Montavon, S.~Lapuschkin, A.~Binder, W.~Samek, and K.-R. M{\"{u}}ller,
  ``Explaining nonlinear classification decisions with deep {T}aylor
  decomposition,'' \emph{Pattern Recognition}, vol.~65, pp. 211--222, 2017.

\bibitem{DBLP:journals/corr/SimonyanZ14a}
K.~Simonyan and A.~Zisserman, ``Very deep convolutional networks for
  large-scale image recognition,'' in \emph{{ICLR}}, 2015.

\bibitem{mva19}
M.~Valleriani, F.~Kr{\"a}utli, M.~Zamani, A.~Tejedor, C.~Sander, M.~Vogl,
  S.~Bertram, G.~Funke, and H.~Kantz, ``The emergence of epistemic communities
  in the sphaera corpus: Mechanisms of knowledge evolution,'' \emph{Journal of
  Historical Network Research}, vol.~3, pp. 50--91, 2019.

\bibitem{Roweis2000}
S.~T. Roweis and L.~K. Saul, ``{Nonlinear Dimensionality Reduction by Locally
  Linear Embedding},'' \emph{Science}, vol. 290, no. 5500, pp. 2323--2326,
  2000.

\bibitem{Coifman2006}
R.~R. Coifman and S.~Lafon, ``Diffusion maps,'' \emph{Applied and Computational
  Harmonic Analysis}, vol.~21, no.~1, pp. 5--30, Jul. 2006.

\bibitem{DBLP:conf/eccv/ZeilerF14}
M.~D. Zeiler and R.~Fergus, ``Visualizing and understanding convolutional
  networks,'' in \emph{{ECCV} {(1)}}, ser. Lecture Notes in Computer Science,
  vol. 8689.\hskip 1em plus 0.5em minus 0.4em\relax Springer, 2014, pp.
  818--833.

\bibitem{DBLP:conf/iclr/ZintgrafCAW17}
L.~M. Zintgraf, T.~S. Cohen, T.~Adel, and M.~Welling, ``Visualizing deep neural
  network decisions: Prediction difference analysis,'' in \emph{{ICLR}
  (Poster)}.\hskip 1em plus 0.5em minus 0.4em\relax OpenReview.net, 2017.

\bibitem{DBLP:conf/nips/LundbergL17}
S.~M. Lundberg and S.~Lee, ``A unified approach to interpreting model
  predictions,'' in \emph{{NIPS}}, 2017, pp. 4765--4774.

\bibitem{DBLP:journals/corr/SimonyanVZ13}
K.~Simonyan, A.~Vedaldi, and A.~Zisserman, ``Deep inside convolutional
  networks: Visualising image classification models and saliency maps,'' in
  \emph{{ICLR} (Workshop Poster)}, 2014.

\bibitem{DBLP:journals/corr/SmilkovTKVW17}
D.~Smilkov, N.~Thorat, B.~Kim, F.~B. Vi{\'{e}}gas, and M.~Wattenberg,
  ``Smooth{G}rad: removing noise by adding noise,'' \emph{CoRR}, vol.
  abs/1706.03825, 2017.

\bibitem{DBLP:conf/icml/SundararajanTY17}
M.~Sundararajan, A.~Taly, and Q.~Yan, ``Axiomatic attribution for deep
  networks,'' in \emph{{ICML}}, ser. Proceedings of Machine Learning Research,
  vol.~70.\hskip 1em plus 0.5em minus 0.4em\relax {PMLR}, 2017, pp. 3319--3328.

\bibitem{Kauffmann20}
J.~Kauffmann, K.-R. M{\"u}ller, and G.~Montavon, ``Towards explaining
  anomalies: A deep {T}aylor decomposition of one-class models,'' \emph{Pattern
  Recognition}, p. 107198, 2020.

\bibitem{DBLP:conf/icdm/MicenkovaNDA13}
B.~Micenkov{\'{a}}, R.~T. Ng, X.~Dang, and I.~Assent, ``Explaining outliers by
  subspace separability,'' in \emph{{ICDM}}.\hskip 1em plus 0.5em minus
  0.4em\relax {IEEE} Computer Society, 2013, pp. 518--527.

\bibitem{Kauffmann19}
J.~Kauffmann, M.~Esders, G.~Montavon, W.~Samek, and K.-R. M{\"{u}}ller, ``From
  clustering to cluster explanations via neural networks,'' \emph{CoRR}, vol.
  abs/1906.07633, 2019.

\bibitem{DBLP:conf/iclr/TsangC018}
M.~Tsang, D.~Cheng, and Y.~Liu, ``Detecting statistical interactions from
  neural network weights,'' in \emph{{ICLR} (Poster)}.\hskip 1em plus 0.5em
  minus 0.4em\relax OpenReview.net, 2018.

\bibitem{kaski-pairwise}
T.~Cui, P.~Marttinen, and S.~Kaski, ``Recovering pairwise interactions using
  neural networks,'' \emph{CoRR}, vol. abs/1901.08361, 2019.

\bibitem{leupold2017second}
S.~Leupold, ``Second-order {T}aylor decomposition for explaining spatial
  transformation of images,'' Master's thesis, Technische Universit{\"at}
  Berlin, 2017.

\bibitem{DBLP:conf/kdd/CaruanaLGKSE15}
R.~Caruana, Y.~Lou, J.~Gehrke, P.~Koch, M.~Sturm, and N.~Elhadad,
  ``Intelligible models for healthcare: Predicting pneumonia risk and hospital
  30-day readmission,'' in \emph{{KDD}}.\hskip 1em plus 0.5em minus 0.4em\relax
  {ACM}, 2015, pp. 1721--1730.

\bibitem{DBLP:journals/corr/abs-2002-04138}
J.~D. Janizek, P.~Sturmfels, and S.~Lee, ``Explaining explanations: Axiomatic
  feature interactions for deep networks,'' \emph{CoRR}, vol. abs/2002.04138,
  2020.

\bibitem{Watkins99dynamicalignment}
C.~Watkins, ``Dynamic alignment kernels,'' in \emph{Advances in Large Margin
  Classifiers}.\hskip 1em plus 0.5em minus 0.4em\relax MIT Press, 1999, pp.
  39--50.

\bibitem{DBLP:journals/neco/TsudaKRSM02}
K.~Tsuda, M.~Kawanabe, G.~R{\"{a}}tsch, S.~Sonnenburg, and K.-R. M{\"{u}}ller,
  ``A new discriminative kernel from probabilistic models,'' \emph{Neural
  Computation}, vol.~14, no.~10, pp. 2397--2414, 2002.

\bibitem{DBLP:journals/sigkdd/Gartner03}
T.~G{\"{a}}rtner, ``A survey of kernels for structured data,'' \emph{{SIGKDD}
  Explorations}, vol.~5, no.~1, pp. 49--58, 2003.

\bibitem{DBLP:conf/nips/BromleyGLSS93}
J.~Bromley, I.~Guyon, Y.~LeCun, E.~S{\"{a}}ckinger, and R.~Shah, ``Signature
  verification using a siamese time delay neural network,'' in
  \emph{{NIPS}}.\hskip 1em plus 0.5em minus 0.4em\relax Morgan Kaufmann, 1993,
  pp. 737--744.

\bibitem{DBLP:conf/cvpr/ChopraHL05}
S.~Chopra, R.~Hadsell, and Y.~LeCun, ``Learning a similarity metric
  discriminatively, with application to face verification,'' in \emph{{CVPR}
  {(1)}}.\hskip 1em plus 0.5em minus 0.4em\relax {IEEE} Computer Society, 2005,
  pp. 539--546.

\bibitem{DBLP:conf/cvpr/WangSLRWPCW14}
J.~Wang, Y.~Song, T.~Leung, C.~Rosenberg, J.~Wang, J.~Philbin, B.~Chen, and
  Y.~Wu, ``Learning fine-grained image similarity with deep ranking,'' in
  \emph{{CVPR}}.\hskip 1em plus 0.5em minus 0.4em\relax {IEEE} Computer
  Society, 2014, pp. 1386--1393.

\bibitem{DBLP:conf/simbad/HofferA15}
E.~Hoffer and N.~Ailon, ``Deep metric learning using triplet network,'' in
  \emph{{SIMBAD}}, ser. Lecture Notes in Computer Science, vol. 9370.\hskip 1em
  plus 0.5em minus 0.4em\relax Springer, 2015, pp. 84--92.

\bibitem{DBLP:conf/eccv/SeguinSdK16}
B.~Seguin, C.~Striolo, I.~diLenardo, and F.~Kaplan, ``Visual link retrieval in
  a database of paintings,'' in \emph{{ECCV} Workshops {(1)}}, ser. Lecture
  Notes in Computer Science, vol. 9913.\hskip 1em plus 0.5em minus 0.4em\relax
  Springer, 2016, pp. 753--767.

\bibitem{DBLP:conf/www/HeLZNHC17}
X.~He, L.~Liao, H.~Zhang, L.~Nie, X.~Hu, and T.~Chua, ``Neural collaborative
  filtering,'' in \emph{{WWW}}.\hskip 1em plus 0.5em minus 0.4em\relax {ACM},
  2017, pp. 173--182.

\bibitem{DBLP:journals/neco/MemisevicH10}
R.~Memisevic and G.~E. Hinton, ``Learning to represent spatial transformations
  with factored higher-order {B}oltzmann machines,'' \emph{Neural Computation},
  vol.~22, no.~6, pp. 1473--1492, 2010.

\bibitem{Tzompanaki2012}
K.~Tzompanaki and M.~Doerr, ``A new framework for querying semantic networks.''
  in \emph{Proceedings of Museums and the Web 2012: the international
  conference for culture and heritage on-line}, 2012.

\bibitem{DBLP:journals/jmlr/GlorotBB11}
X.~Glorot, A.~Bordes, and Y.~Bengio, ``Deep sparse rectifier neural networks,''
  in \emph{{AISTATS}}, ser. {JMLR} Proceedings, vol.~15.\hskip 1em plus 0.5em
  minus 0.4em\relax JMLR.org, 2011, pp. 315--323.

\bibitem{DBLP:conf/cvpr/HeZRS16}
K.~He, X.~Zhang, S.~Ren, and J.~Sun, ``Deep residual learning for image
  recognition,'' in \emph{{CVPR}}.\hskip 1em plus 0.5em minus 0.4em\relax
  {IEEE} Computer Society, 2016, pp. 770--778.

\bibitem{DBLP:journals/corr/ShrikumarGSK16}
A.~Shrikumar, P.~Greenside, A.~Shcherbina, and A.~Kundaje, ``Not just a black
  box: Learning important features through propagating activation
  differences,'' \emph{CoRR}, vol. abs/1605.01713, 2016.

\bibitem{DBLP:conf/iclr/AnconaCO018}
M.~Ancona, E.~Ceolini, C.~{\"{O}}ztireli, and M.~Gross, ``Towards better
  understanding of gradient-based attribution methods for deep neural
  networks,'' in \emph{{ICLR} (Poster)}.\hskip 1em plus 0.5em minus 0.4em\relax
  OpenReview.net, 2018.

\bibitem{axioms}
G.~Montavon, ``Gradient-based vs. propagation-based explanations: An axiomatic
  comparison,'' in \emph{Explainable {AI}}, ser. Lecture Notes in Computer
  Science.\hskip 1em plus 0.5em minus 0.4em\relax Springer, 2019, vol. 11700,
  pp. 253--265.

\bibitem{DBLP:conf/icml/BalduzziFLLMM17}
D.~Balduzzi, M.~Frean, L.~Leary, J.~P. Lewis, K.~W. Ma, and B.~McWilliams,
  ``The shattered gradients problem: If resnets are the answer, then what is
  the question?'' in \emph{{ICML}}, ser. Proceedings of Machine Learning
  Research, vol.~70.\hskip 1em plus 0.5em minus 0.4em\relax {PMLR}, 2017, pp.
  342--350.

\bibitem{lrpoverview}
G.~Montavon, A.~Binder, S.~Lapuschkin, W.~Samek, and K.-R. M{\"{u}}ller,
  ``Layer-wise relevance propagation: An overview,'' in \emph{Explainable
  {AI}}, ser. Lecture Notes in Computer Science.\hskip 1em plus 0.5em minus
  0.4em\relax Springer, 2019, vol. 11700, pp. 193--209.

\bibitem{DBLP:journals/corr/abs-1911-09017}
H.~Zhang, J.~Chen, H.~Xue, and Q.~Zhang, ``Towards a unified evaluation of
  explanation methods without ground truth,'' \emph{CoRR}, vol. abs/1911.09017,
  2019.

\bibitem{lapuschkin2017faces}
S.~Lapuschkin, A.~Binder, K.-R. M{\"u}ller, and W.~Samek, ``Understanding and
  comparing deep neural networks for age and gender classification,'' in
  \emph{{IEEE} International Conference on Computer Vision Workshops}, 2017,
  pp. 1629--1638.

\bibitem{pascal-voc-2007}
M.~Everingham, L.~Van~Gool, C.~K.~I. Williams, J.~Winn, and A.~Zisserman, ``The
  {PASCAL} {V}isual {O}bject {C}lasses {C}hallenge 2007 {(VOC2007)}
  {R}esults,''
  http://www.pascal-network.org/challenges/VOC/voc2007/workshop/index.html.

\bibitem{DBLP:journals/ml/Caruana97}
R.~Caruana, ``Multitask learning,'' \emph{Machine Learning}, vol.~28, no.~1,
  pp. 41--75, 1997.

\bibitem{DBLP:conf/cvpr/OquabBLS14}
M.~Oquab, L.~Bottou, I.~Laptev, and J.~Sivic, ``Learning and transferring
  mid-level image representations using convolutional neural networks,'' in
  \emph{{CVPR}}.\hskip 1em plus 0.5em minus 0.4em\relax {IEEE} Computer
  Society, 2014, pp. 1717--1724.

\bibitem{DBLP:conf/kdd/ZhangLZSKYJ15}
W.~Zhang, R.~Li, T.~Zeng, Q.~Sun, S.~Kumar, J.~Ye, and S.~Ji, ``Deep model
  based transfer and multi-task learning for biological image analysis,'' in
  \emph{{KDD}}.\hskip 1em plus 0.5em minus 0.4em\relax {ACM}, 2015, pp.
  1475--1484.

\bibitem{DBLP:journals/mia/LitjensKBSCGLGS17}
G.~J.~S. Litjens, T.~Kooi, B.~E. Bejnordi, A.~A.~A. Setio, F.~Ciompi,
  M.~Ghafoorian, J.~A. W.~M. van~der Laak, B.~van Ginneken, and C.~I.
  S{\'{a}}nchez, ``A survey on deep learning in medical image analysis,''
  \emph{Medical Image Analysis}, vol.~42, pp. 60--88, 2017.

\bibitem{DBLP:journals/cacie/GaoM18}
Y.~Gao and K.~M. Mosalam, ``Deep transfer learning for image-based structural
  damage recognition,'' \emph{Comp.-Aided Civil and Infrastruct. Engineering},
  vol.~33, no.~9, pp. 748--768, 2018.

\bibitem{DBLP:conf/micai/LencK15}
L.~Lenc and P.~Kr{\'{a}}l, ``Unconstrained facial images: Database for face
  recognition under real-world conditions,'' in \emph{{MICAI} {(2)}}, ser.
  Lecture Notes in Computer Science, vol. 9414.\hskip 1em plus 0.5em minus
  0.4em\relax Springer, 2015, pp. 349--361.

\bibitem{LFWTech}
G.~B. Huang, M.~Ramesh, T.~Berg, and E.~Learned-Miller, ``Labeled faces in the
  wild: A database for studying face recognition in unconstrained
  environments,'' University of Massachusetts, Amherst, Tech. Rep. 07-49,
  October 2007.

\bibitem{mva20}
M.~Valleriani, ``Prolegomena to the study of early modern commentators on
  {J}ohannes de {S}acrobosco's tractatus de sphaera,'' in \emph{De sphaera of
  Johannes de Sacrobosco in the Early Modern Period: The Authors of the
  Commentaries}.\hskip 1em plus 0.5em minus 0.4em\relax Springer Nature, 2019,
  pp. 1--23.

\bibitem{DBLP:conf/cvpr/SunWT14}
Y.~Sun, X.~Wang, and X.~Tang, ``Deep learning face representation from
  predicting 10, 000 classes,'' in \emph{{CVPR}}.\hskip 1em plus 0.5em minus
  0.4em\relax {IEEE} Computer Society, 2014, pp. 1891--1898.

\bibitem{DBLP:journals/lalc/KrautliV18}
F.~Kr{\"{a}}utli and M.~Valleriani, ``Corpus{T}racer: {A} {CIDOC} database for
  tracing knowledge networks,'' \emph{{DSH}}, vol.~33, no.~2, pp. 336--346,
  2018.

\bibitem{Lang18}
S.~Lang and B.~Ommer, ``{Attesting similarity: Supporting the organization and
  study of art image collections with computer vision},'' \emph{Digital
  Scholarship in the Humanities}, vol.~33, no.~4, pp. 845--856, 04 2018.

\bibitem{DBLP:journals/pami/BrunaM13}
J.~Bruna and S.~Mallat, ``Invariant scattering convolution networks,''
  \emph{{IEEE} Trans. Pattern Anal. Mach. Intell.}, vol.~35, no.~8, pp.
  1872--1886, 2013.

\bibitem{Chmiela2018}
S.~Chmiela, H.~E. Sauceda, K.-R. M\"{u}ller, and A.~Tkatchenko, ``Towards exact
  molecular dynamics simulations with machine-learned force fields,''
  \emph{Nature Communications}, vol.~9, no.~1, Sep. 2018.

\bibitem{Anselmi2016}
F.~Anselmi, L.~Rosasco, and T.~Poggio, ``On invariance and selectivity in
  representation learning,'' \emph{Information and Inference}, vol.~5, no.~2,
  pp. 134--158, May 2016.

\bibitem{DBLP:conf/nips/GoodfellowLSLN09}
I.~J. Goodfellow, Q.~V. Le, A.~M. Saxe, H.~Lee, and A.~Y. Ng, ``Measuring
  invariances in deep networks,'' in \emph{{NIPS}}.\hskip 1em plus 0.5em minus
  0.4em\relax Curran Associates, Inc., 2009, pp. 646--654.

\bibitem{Rodriguez2008ActionMA}
M.~D. Rodriguez, J.~Ahmed, and M.~Shah, ``Action {MACH} a spatio-temporal
  maximum average correlation height filter for action recognition,''
  \emph{2008 IEEE Conference on Computer Vision and Pattern Recognition}, pp.
  1--8, 2008.

\bibitem{ucfsports2014}
K.~Soomro and A.~Zamir, ``Action recognition in realistic sports videos,''
  \emph{Advances in Computer Vision and Pattern Recognition}, vol.~71, pp.
  181--208, 01 2014.

\bibitem{DBLP:conf/iccv/DosovitskiyFIHH15}
A.~Dosovitskiy, P.~Fischer, E.~Ilg, P.~H{\"{a}}usser, C.~Hazirbas, V.~Golkov,
  P.~van~der Smagt, D.~Cremers, and T.~Brox, ``Flownet: Learning optical flow
  with convolutional networks,'' in \emph{{ICCV}}.\hskip 1em plus 0.5em minus
  0.4em\relax {IEEE} Computer Society, 2015, pp. 2758--2766.

\bibitem{husson14}
M.~Husson, ``Remarks on two dimensional array tables in latin astronomy: a case
  study in layout transmission,'' \emph{Suhayl. Journal for the History of the
  Exact and Natural Sciences in Islamic Civilisation}, vol.~13, pp. 103--117,
  2014.

\bibitem{DBLP:conf/icdar/SchreiberAWDA17}
S.~Schreiber, S.~Agne, I.~Wolf, A.~Dengel, and S.~Ahmed, ``Deep{D}e{SRT}: Deep
  learning for detection and structure recognition of tables in document
  images,'' in \emph{{ICDAR}}.\hskip 1em plus 0.5em minus 0.4em\relax {IEEE},
  2017, pp. 1162--1167.

\end{thebibliography}


\begin{thebibliography}{1}
\providecommand{\url}[1]{#1}
\csname url@samestyle\endcsname
\providecommand{\newblock}{\relax}
\providecommand{\bibinfo}[2]{#2}
\providecommand{\BIBentrySTDinterwordspacing}{\spaceskip=0pt\relax}
\providecommand{\BIBentryALTinterwordstretchfactor}{4}
\providecommand{\BIBentryALTinterwordspacing}{\spaceskip=\fontdimen2\font plus
\BIBentryALTinterwordstretchfactor\fontdimen3\font minus
  \fontdimen4\font\relax}
\providecommand{\BIBforeignlanguage}[2]{{%
\expandafter\ifx\csname l@#1\endcsname\relax
\typeout{** WARNING: IEEEtran.bst: No hyphenation pattern has been}%
\typeout{** loaded for the language `#1'. Using the pattern for}%
\typeout{** the default language instead.}%
\else
\language=\csname l@#1\endcsname
\fi
#2}}
\providecommand{\BIBdecl}{\relax}
\BIBdecl

\bibitem{DBLP:journals/pr/MontavonLBSM17}
G.~Montavon, S.~Lapuschkin, A.~Binder, W.~Samek, and K.-R. M{\"{u}}ller,
  ``Explaining nonlinear classification decisions with deep {T}aylor
  decomposition,'' \emph{Pattern Recognition}, vol.~65, pp. 211--222, 2017.

\bibitem{lrp}
S.~Bach, A.~Binder, G.~Montavon, F.~Klauschen, K.-R. M{\"u}ller, and W.~Samek,
  ``On pixel-wise explanations for non-linear classifier decisions by
  layer-wise relevance propagation,'' \emph{PLoS ONE}, vol.~10, no.~7, p.
  e0130140, 07 2015.

\bibitem{lrpoverview}
G.~Montavon, A.~Binder, S.~Lapuschkin, W.~Samek, and K.-R. M{\"{u}}ller,
  ``Layer-wise relevance propagation: An overview,'' in \emph{Explainable
  {AI}}, ser. Lecture Notes in Computer Science.\hskip 1em plus 0.5em minus
  0.4em\relax Springer, 2019, vol. 11700, pp. 193--209.

\bibitem{DBLP:journals/corr/ShrikumarGSK16}
A.~Shrikumar, P.~Greenside, A.~Shcherbina, and A.~Kundaje, ``Not just a black
  box: Learning important features through propagating activation
  differences,'' \emph{CoRR}, vol. abs/1605.01713, 2016.

\bibitem{DBLP:series/lncs/LeCunBOM12}
Y.~LeCun, L.~Bottou, G.~B. Orr, and K.-R. M{\"{u}}ller, ``Efficient backprop,''
  in \emph{Neural Networks: Tricks of the Trade (2nd ed.)}, ser. Lecture Notes
  in Computer Science.\hskip 1em plus 0.5em minus 0.4em\relax Springer, 2012,
  vol. 7700, pp. 9--48.

\end{thebibliography}

\end{document}


\title{
Building and Interpreting Deep Similarity Models\\{\large \sc \sffamily (Supplementary Material)}
}

\author{Oliver Eberle, Jochen B\"uttner, Florian Kr\"autli, Klaus-Robert M\"uller, Matteo Valleriani, Gr\'egoire Montavon}

\maketitle

\newcommand{\z}{\boldsymbol{z}}
\newcommand{\x}{\boldsymbol{x}}
\newcommand{\ba}{\boldsymbol{a}}
\newcommand{\rba}{\widetilde{\ba}}
\newcommand{\bx}{\boldsymbol{x}}
\newcommand{\rx}{\widetilde{x}}
\newcommand{\rbx}{\widetilde{\bx}}
\newcommand{\w}{\boldsymbol{w}}

\appendices

\noindent In this Supplement, we give proofs and derivations for the `\HessProd{}' baseline and for our proposed \acronym{} method. We also give details on the procedure we use in the paper to render \acronym{} explanations on image data.

\section{`\HessProd{}' Baseline}

The `\HessProd{}' (HP) baseline we consider in this paper applies to similarity models of the type:
$$
y(\x,\x') = \langle \phi(\x),\phi(\x') \rangle
$$
a dot product on a feature map $\phi\colon\mathbb{R}^d \to \mathbb{R}^h$ satisfying first-order positive homogeneity i.e.\ $\forall_{\x},\forall_{t>0}:~\phi(t\x) = t\phi (\x)$.

\subsection{Derivation of HP}

We derive \HessProd{} as the result of a Taylor expansion of the similarity model at the root point $(\rbx,\rbx') = (\varepsilon \kern 0.01em \x,\varepsilon \kern 0.01em \x')$ with $\varepsilon$ almost zero. For the zero-order term, we get:
\begin{align*}
y(\rbx,\rbx') &= 0
\end{align*}
Let $\nabla$ and $\nabla'$ be the gradient operators with respect to the features forming $\x$ and $\x'$ respectively. First-order terms associated to the features of $\x$ are given by:
\begin{align*}
R_i &= [\nabla y(\rbx,\rbx')]_i \cdot (x_i - \rx_i)\\
&= \textstyle \big[\nabla \sum_m \phi_m(\rbx)\phi_m(\rbx')\big]_i \cdot x_i\\
&= \textstyle \big[\sum_m (\nabla \phi_m(\rbx))\cdot \phi_m(\rbx')\big]_i \cdot x_i\\
&= \textstyle \big[\sum_m (\nabla \phi_m(\rbx))\cdot \phi_m(\boldsymbol{0})\big]_i \cdot x_i\\
&= 0
\end{align*}
In a similar way, for the features of $\x'$, we get $R_{i'} = 0$. To extract the interaction terms $R_{ii'}$, we first show that $\forall_{t>0}:~\nabla \phi_m(t\x) = \nabla \phi_m(\x)$:
\begin{align*}
\nabla \phi_m(t\x) = t^{-1} \frac{\partial}{\partial \x} \phi_m(t\x) = t^{-1}\frac{\partial}{\partial \x} t\phi_m(\x) = \nabla\phi_m(\x)
\end{align*}
Then, we develop the interaction terms of the Taylor expansion as:
\begin{align}
R_{ii'}
&= [\nabla^2 y(\rbx,\rbx')]_{ii'} \cdot (x_i - \rx_i) \cdot (x'_{i'} - \rx'_{i'})\nonumber\\
&= \textstyle \big[ \sum_m \nabla^2 \phi_m(\rbx)\phi_m(\rbx')\big]_{ii'} \cdot x_i x'_{i'}\nonumber\\
&= \textstyle \big[ \sum_m \nabla \nabla' \phi_m(\rbx)\phi_m(\rbx')\big]_{ii'} \cdot x_i x'_{i'}\nonumber\\
&= \textstyle \big[ \sum_m (\nabla \phi_m(\rbx)) \otimes (\nabla' \phi_m(\rbx'))\big]_{ii'}  \cdot x_i x'_{i'}\nonumber
\intertext{where $\otimes$ denotes the outer product, we apply the property shown above ($\nabla \phi_m(t\x) = \nabla \phi_m(\x)$) to get}
&= \textstyle \big[ \sum_m (\nabla \phi_m(\x)) \otimes (\nabla' \phi_m(\x'))\big]_{ii'}  \cdot x_i x'_{i'}\label{eq:outer}
\intertext{and finally, we apply the steps in reverse order}
&= \textstyle \big[\sum_m \nabla \nabla' \phi_m(\bx)\phi_m(\bx')\big]_{ii'} \cdot x_i x'_{i'}\nonumber\\
&= \textstyle \big[\sum_m \nabla^2 \phi_m(\bx)\phi_m(\bx')\big]_{ii'} \cdot x_i x'_{i'}\nonumber\\
&= [\nabla^2 y(\x,\x')]_{ii'} \cdot x_i x'_{i'}.\nonumber
\end{align}
The last line corresponds to the HP baseline.

\subsection{Conservation of HP}

We show that `\HessProd{}' sums to the similarity score, and thus constitutes an explanation that is conservative. For this, we can first show that $\x^\top \nabla \phi_m(\x) = \phi_m(\x)$:
\begin{align*}
\x^\top \nabla \phi_m(t\x) = \frac{\partial}{\partial t} \phi_m(t\x) = \frac{\partial}{\partial t} t\phi_m(\x) = \phi_m(\x)
\end{align*}
Choosing $t=1$ completes the proof. (This result is known as Euler's homogeneous function theorem.)

\smallskip

\noindent Starting from Eq.\ \eqref{eq:outer}, we then write:
\begin{align*}
\textstyle \sum_{ii'} R_{ii'}
&= \textstyle \sum_{ii'} \big[ \sum_m (\nabla \phi_m(\x)) \otimes (\nabla' \phi_m(\x'))\big]_{ii'}  \cdot x_i x'_{i'}\\
&= \textstyle \sum_m \sum_{i} x_i [\nabla \phi_m(\x)]_i \cdot \sum_{i'} x'_{i'} [\nabla' \phi_m(\x')]_{i'} \\
&= \textstyle \sum_m \x^\top \nabla \phi_m(\x) \cdot \x'^\top \nabla' \phi_m(\x')\\
&= \textstyle \sum_m \phi_m(\x) \cdot \phi_m(\x')\\
&= y(\x,\x')
\end{align*}
which shows that the explanation is conservative.

\subsection{`\GradInput{}' Formulation of HP}

We show that `\HessProd{}' can be rewritten as $2 \times h$ `\GradInput{}' (GI) computations. Starting from Eq.\ \eqref{eq:outer}, we get:
\begin{align*}
R_{ii'} &= \textstyle \big[ \sum_m (\nabla \phi_m(\x)) \otimes (\nabla' \phi_m(\x'))\big]_{ii'}  \cdot x_i x'_{i'}\\
&= \textstyle \big[ \sum_m (\nabla \phi_m(\x) \odot \x) \otimes (\nabla' \phi_m(\x')  \odot \x')]_{ii'}\\
&=\textstyle \big[ \sum_m  \text{GI}(\phi_m,\x) \otimes \text{GI} (\phi_m,\x') \big]_{ii'}
\end{align*}
Therefore, scores $R_{ii'}$ produced by HP are the elements of a sum of outer products of GI computations.

\section{Derivation of \acronym{}}

The deep Taylor decomposition \cite{DBLP:journals/pr/MontavonLBSM17} (DTD) framework we use to derive \acronym{} propagation rules assumes that relevance propagated up to a certain layer can be modeled as
$$
\widehat{R}_{kk'}(\ba) = a_k a_{k'} c_{kk'}
$$
i.e.\ a product of activations in the two branches of the similarity computation, multiplied by a term $c_{kk'}$ assumed to be constant and set in a way that $\widehat{R}_{kk'}(\ba) = R_{kk'}$. DTD seeks to propagate the modeled relevance to the layer below by identifying the terms of a Taylor expansion. In the following, we distinguish between (1) linear/ReLU layers, and (2) positively homogeneous layers (e.g.\ min- or max-pooling).

\subsection{Linear/ReLU Layers}

These layers produce output activations of the type
\begin{align*}
a_k &= \textstyle \big(\sum_{j} a_j w_{jk}\big)^+\\
a_{k'} &= \textstyle \big(\sum_{j'} a_{j'} w_{j'k'}\big)^+
\end{align*}
where the weighted sum can be either a dense layer, or a convolution. The relevance model can be written as:
\begin{align*}
\widehat{R}_{kk'}(\ba)  &= a_k a_{k'} c_{kk'}\\
&= \textstyle \big(\sum_j a_j w_{jk}\big)^+ \big( \sum_{j'} a_{j'} w_{j'k'}\big)^+ c_{kk'}
\end{align*}
When neurons $a_k$ and $a_{k'}$ are jointly activated (i.e.\ $a_k,a_{k'} > 0$), a second-order Taylor expansion of $R_{kk'}$ at some reference point $\rba$ is given by:
\begin{align*}
\widehat{R}_{kk'}(\ba) &= \textstyle  \big(\sum_{j} \widetilde{a}_{j} w_{jk}\big) \big(\sum_{j'} \widetilde{a}_{j'} w_{j'k'}\big) c_{kk'}\\
& \quad \textstyle + \sum_{j} (a_j - \widetilde{a}_j) w_{jk} \big(\sum_{j'} \widetilde{a}_{j'} w_{j'k'}\big) c_{kk'}\\
& \quad \quad \textstyle +  \sum_{j'} \big(\sum_{j} \widetilde{a}_{j} w_{jk}\big) (a_{j'} - \widetilde{a}_{j'}) w_{j'k'} c_{kk'}\\
& \quad \quad \quad \textstyle + \sum_{jj'} \textstyle (a_j - \widetilde{a}_j) w_{jk} (a_{j'} - \widetilde{a}_{j'}) w_{j'k'} c_{kk'}
\end{align*}
\acronym{} chooses the reference point $\rba$ to be subject to the following two constraints:
\begin{enumerate}
\item very close to the ReLU hinges of neurons $k$ and $k'$ (but still on the activated domain)
\item on the plane $\{\rba(t,t') |~ t,t' \in \mathbb{R}\}$
where
\begin{align*}
[\rba(t,t')]_j &= a_j - t a_j \cdot (1 + \gamma \cdot 1_{w_{jk} > 0})\\
[\rba(t,t')]_{j'} &= a_{j'} - t' a_{j'} \cdot (1 + \gamma \cdot 1_{w_{j'k'} > 0})
\end{align*}
with $\gamma$ a hyperparameter.
\end{enumerate}
We now analyze the different terms of the expansion at this reference point.
\begin{itemize}
\item The zero-order term is zero.
\item The first-order terms are also zero because the reference point is chosen at the {\em intersection} of the ReLU hinges of neurons $k$ and $k'$, hence the non-differentiated term is zero.
\item The interaction terms are given by:
\begin{align*}
R_{jj' \leftarrow kk'} &= t a_j (1 + \gamma 1_{w_{jk}>0}) \\
& \qquad \cdot~t' a_{j'} (1 + \gamma 1_{w_{j'k'}>0})\\
& \qquad \qquad \cdot~w_{jk} w_{j'k'} c_{kk'}\\
&= tt' a_j a_{j'} \rho(w_{jk}) \rho(w_{j'k'}) c_{kk'}
\end{align*}
where $\rho(w_{jk}) = w_{jk} + \gamma w_{jk}^+$ and where the product of parameters $tt'$ must still be resolved.
\end{itemize}
Because we expand a bilinear form, and because zero-order and first-order terms are zero, the constraint $\sum_{jj'} R_{jj' \leftarrow kk'} = R_{kk'}$ must be satisfied. This constraint allows us to resolve the product $tt'$, leading to the following closed-form expression for the interaction terms:
\begin{align*}
R_{jj' \leftarrow kk'} &= \frac{a_j a_{j'} \rho(w_{jk}) \rho(w_{j'k'})}{\sum_{jj'} a_j a_{j'} \rho(w_{jk}) \rho(w_{j'k'})} R_{kk'}
\end{align*}
This propagation rule is also consistent with the case where $a_k$ or $a_{k'}$ are zero and where no relevance needs to be redistributed. Aggregate relevance scores for the layer below are obtained by summing over neurons in the higher-layer:
\begin{align}
\textstyle 
R_{jj'} &= \textstyle \sum_{kk'} R_{jj' \leftarrow kk'}\nonumber\\
&=\sum_{kk'} \frac{a_j a_{j'} \rho(w_{jk}) \rho(w_{j'k'})}{\sum_{jj'} a_j a_{j'} \rho(w_{jk}) \rho(w_{j'k'})} R_{kk'}
\label{eq:Rjj}
\end{align}
This last equation is the propagation rule used by \acronym{} to propagate relevance in linear/ReLU layers.

\subsection{Positively Homogeneous Layers}

When $a_k$ and $a_{k'}$ are positively homogeneous functions of their input activations (e.g.\ min- and max-pooling layers), the relevance model can be expressed in terms of the Hessian:
\begin{align*}
\widehat{R}_{kk'}(\ba) &= a_k a_{k'} c_{kk'}\\
&= \textstyle 
\big(\sum_j a_j [\nabla a_k]_j\big)
\big(\sum_{j'} a_{j'} [\nabla a_{k'}]_{j'}\big)
c_{kk'}\\
&= \textstyle \sum_{jj'} a_j a_{j'} [\nabla^2 a_k a_{k'}]_{jj'} c_{kk'}
\end{align*}
The last form can also be interpreted as the interaction terms of a Taylor expansion of $\widehat{R}_{kk'}$ at $\rba = \varepsilon \kern 0.01em \ba$ with $\varepsilon$ almost zero. Zero-order and first-order terms of the expansion vanish, and interaction terms can be rewritten in a propagation-like manner as:
$$
R_{jj' \leftarrow kk'} = \frac{a_j a_{j'} [\nabla^2 a_k a_{k'}]_{jj'}}{\sum_{jj'} a_j a_{j'} [\nabla^2 a_k a_{k'}]_{jj'}} R_{kk'},
$$
and finally,
\begin{align}
R_{jj'} &= \sum_{kk'} \frac{a_j a_{j'} [\nabla^2 a_k a_{k'}]_{jj'}}{\sum_{jj'} a_j a_{j'} [\nabla^2 a_k a_{k'}]_{jj'}} R_{kk'},
\label{eq:Rjjother}
\end{align}
which is the \acronym{} propagation rule we use in these layers.

\section{Factorization of \acronym{}}
\label{appendix:bilrp-factorization}

In this appendix, we show how the propagation rules in Equations \eqref{eq:Rjj} and \eqref{eq:Rjjother} can be factorized to be expressed as compositions of standard LRP \cite{lrp} propagation rules. In the top layer, the dot product similarity can be written as:
\begin{align*}
\textstyle y = \sum_{kk'} a_k a_{k'} 1_{\text{id}(k) = \text{id}(k')}
\end{align*}
where `id' is a function returning the neuron index in its respective branch (a number from $1$ to $h$). Relevance scores can be identified and developed as:
\begin{align*}
R_{kk'} &= a_k a_{k'} 1_{\text{id}(k) = \text{id}(k')}\\
&= \textstyle a_k a_{k'} \sum_{m=1}^h 1_{\text{id}(k) = m} 1_{\text{id}(k') = m}\\
&= \textstyle \sum_{m=1}^h {\underbrace{a_k 1_{\text{id}(k)=m}}_{R_{km}}} \cdot {\underbrace{ a_{k'} 1_{\text{id}(k')=m}}_{R_{k'm}}}
\end{align*}
where we have extracted the desired factor structure. We now apply an inductive argument: Assume that at some layer, $R_{kk'}$ factorizes as $R_{kk'} =\sum_{m=1}^h R_{km} R_{k'm}$. We can show that the same holds in the layer below, in particular, Eq. \eqref{eq:Rjj} can be rewritten as:
\begin{align*}
R_{jj'} &= \sum_{kk'}\frac{a_j a_{j'} \rho(w_{jk}) \rho(w_{j'k'})}{\sum_{jj'} a_j a_{j'} \rho(w_{jk}) \rho(w_{j'k'})} \sum_{m=1}^h R_{km} R_{k'm}\\
&= \sum_{m=1}^h \sum_{kk'}\frac{a_j \rho(w_{jk}) a_{j'} \rho(w_{j'k'})}{\sum_{j} a_j \rho(w_{jk}) \sum_{j'} a_{j'} \rho(w_{j'k'})} R_{km} R_{k'm}\\
&= \footnotesize \sum_{m=1}^h \Big(\underbrace{\sum_{k}\frac{a_j \rho(w_{jk})}{\sum_{j} a_j \rho(w_{jk})} R_{km}\!}_{R_{jm}}\Big)\!\cdot\! \Big( \underbrace{\sum_{k'}\frac{a_{j'} \rho(w_{j'k'})}{\sum_{j'} a_{j'} \rho(w_{j'k'})} R_{k'm}\!}_{R_{j'm}}\Big)
\end{align*}
where we identify a similar factorization. Furthermore, terms of the factorization can be computed using standard LRP rules, here, LRP-$\gamma$ \cite{lrpoverview}. Similarly, Eq.\ \eqref{eq:Rjjother} can be rewritten as:
\begin{align*}
R_{jj'} &= \sum_{kk'}\frac{a_j a_{j'} [\nabla^2 a_k a_{k'}]_{jj'}}{\sum_{jj'} a_j a_{j'} [\nabla^2 a_k a_{k'}]_{jj'}} \sum_{m=1}^h R_{km} R_{k'm}\\
&= \sum_{m=1}^h \sum_{kk'}\frac{a_j [\nabla a_k]_{j} a_{j'} [\nabla a_{k'}]_{j'}}{\sum_{j} a_j [\nabla a_k]_{j} \sum_{j'}a_{j'} [\nabla a_{k'}]_{j'}} R_{km} R_{k'm}\\
&= \footnotesize \sum_{m=1}^h \Big(\underbrace{\sum_{k}\frac{a_j [\nabla a_k]_{j}}{\sum_{j} a_j [\nabla a_k]_{j}} R_{km}\!}_{R_{jm}}\Big)\!\cdot\! \Big( \underbrace{\sum_{k'}\frac{a_{j'} [\nabla a_{k'}]_{j'}}{\sum_{j'} a_{j'} [\nabla a_{k'}]_{j'}} R_{k'm}\!}_{R_{j'm}}\Big)
\end{align*}
which again factorizes into a composition of LRP-type propagation rules.

\section{Theoretical Properties of \acronym{}}

In this appendix, we give proofs for the theoretical properties stated in Section 3.3 of the paper.

\subsection{Conservation of \acronym{}}

An important property of LRP \cite{lrp} is conservation, i.e.\ the relevance scores assigned to the input features sum to the prediction output\footnote{In LRP, exact conservation requires using non-dissipative propagation rules (e.g.\ LRP-0 and LRP-$\gamma$), as well as avoiding contribution of biases (e.g.\ by training a model with biases set to zero).}. Similar results can be obtained for \acronym{}.

\begin{proposition}
For deep rectifier networks with zero biases, \acronym{} is conservative, i.e.\ $\sum_{ii'}R_{ii'} = y(\x,\x')$.
\end{proposition}

\noindent We first show conservation when propagating with Eq.\ \eqref{eq:Rjj} in a linear/ReLU layer:
\begin{align*}
{\textstyle \sum_{jj'} R_{jj'}} &= \sum_{jj'}\sum_{kk'}\frac{a_j a_{j'} \rho(w_{jk}) \rho(w_{j'k'})}{\sum_{jj'} a_j a_{j'} \rho(w_{jk}) \rho(w_{j'k'})} R_{kk'}\\
&= \sum_{kk'}\frac{\sum_{jj'} a_j a_{j'} \rho(w_{jk}) \rho(w_{j'k'})}{\sum_{jj'} a_j a_{j'} \rho(w_{jk}) \rho(w_{j'k'})} R_{kk'} = \textstyle \sum_{kk'}R_{kk'}
\end{align*}
Same conservation property can be shown for the propagation rule in Eq.\ \eqref{eq:Rjjother}. Because these rules are applied repeatedly at each layer, we get the chain of equalities
\begin{align*}
\textstyle
\sum_{ii'} R_{ii'} = \dots = 
\sum_{jj'} R_{jj'} = 
\sum_{kk'} R_{kk'} = \dots = y(\x,\x')
\end{align*}
where we observe that conservation also holds globally.

\subsection{HP as a Special Case of \acronym{}}

A result due to \cite{DBLP:journals/corr/ShrikumarGSK16} is that application of a special case of LRP (referred by \cite{lrpoverview} as LRP-0, or LRP-$\gamma$ with $\gamma=0$) at each layer of the network produces an explanation that is equivalent to \GradInput{}. A similar result can be shown for \acronym{}.

\begin{proposition}
When $\gamma=0$, explanations produced by \acronym{} reduce to those of \HessProd{}.
\end{proposition}

\noindent Rewriting relevance scores as $R_{jj'} = a_j a_{j'} c_{jj'}$ and $R_{kk'} = a_k a_{k'} c_{kk'}$ and observing that for $\gamma=0$, we have $\rho(w_{jk}) = w_{jk}$, the propagation from one layer to another can be written for Eq.\ \eqref{eq:Rjj} as:
\begin{align*}
c_{jj'} &= \sum_{kk'} w_{jk} w_{j'k'} \frac{a_k}{\sum_j a_j w_{jk}}\frac{a_{k'}}{\sum_{j'}a_{j'} w_{j'k'}} c_{kk'}\\
&= \sum_{kk'} w_{jk} w_{j'k'} 1_{a_k>0}1_{a_{k'}>0} c_{kk'}\\
&= \sum_{kk'} [\nabla a_k]_j [\nabla a_{k'}]_{j'} c_{kk'}
\intertext{and similarly for Eq.\ \eqref{eq:Rjjother} as:}
c_{jj'} &= \sum_{kk'} [\nabla^2 a_k a_{k'}]_{jj'} c_{kk'}\\
&= \sum_{kk'} [\nabla a_k]_j [\nabla a_{k'}]_{j'} c_{kk'}
\end{align*}
For the considered class of functions, this relation is equivalent to the formula for propagating second-order derivatives (cf.\ \cite{DBLP:series/lncs/LeCunBOM12}), where $c_{jj'}$ and $c_{kk'}$ denote $[\nabla^2 y]_{jj'}$ and $[\nabla^2 y]_{kk'}$ respectively. Hence, we get at the end of the LRP procedure the quantity $c_{ii'} = [\nabla^2 y]_{ii'}$ and therefore $R_{ii'} = x_i x'_{i'} c_{ii'}$ is equivalent to `\HessProd{}'.
\subsection{Product Approximation in \acronym{}}

We highlight in the following the product structure of relevance scores produced by \acronym{} at each layer. This product structure supports the relevance model used by DTD, from which \acronym{} propagation rules can be derived.

\begin{proposition} The relevance computed by \acronym{} at each layer can be rewritten as $R_{jj'} = a_j a_{j'} c_{jj'}$, where $c_{jj'}$ is locally approximately constant.
\label{prop:model}
\end{proposition}

\noindent In the top layer, we have $c_{kk'} = 1_{\text{id}(k) = \text{id}(k')}$ (cf.\ Appendix \ref{appendix:bilrp-factorization}), which is constant. We apply an inductive argument: Assume that at some layer, $c_{kk'}$ is locally approximately constant, we would like to show that the same holds for $c_{jj'}$ in the layer below.

Relevance scores in Eq.\ \eqref{eq:Rjj} can be rewritten as $R_{jj'} = a_j a_{j'} c_{jj'}$ with:
\begin{align*}
c_{jj'} = \sum_{kk'} \rho(w_{jk}) \rho(w_{j'k'}) \frac{\big(\sum_{j} a_j w_{jk}\big)^{\!+} \big(\sum_{j'} a_{j'} w_{j'k'}\big)^{\!+}\!\!}{\sum_{jj'} a_j a_{j'} \rho(w_{jk}) \rho(w_{j'k'})}\, c_{kk'}.
\end{align*}

\noindent The term $c_{jj'}$ depends on $a_j$ and $a_{j'}$ only through (1) nested sums, which can be seen as diluting the effect of these activations, and (2) the term $c_{kk'}$ which we have assumed as a starting point to be locally approximately constant.

Similarly, for Eq.\ \eqref{eq:Rjjother}, the redistributed relevance can be written in product form, with $\textstyle c_{jj'} = \sum_{kk'} \,[\nabla^2 a_k a_{k'}]_{jj'} c_{kk'}$. This time, $c_{jj'}$ depends on local activations through (1) a combination of a nested sum and a second-order differentiation, with the same diluting effect as above, and (2) the term $c_{kk'}$ which is locally approximately constant.

\smallskip

Overall, in both cases, the weak dependency of $c_{jj'}$ on local activations provides support for treating this term as constant in the relevance model used by DTD.

\section{Coarse-Grained Explanations}
\label{section:bilrp-pooling}

When the input has $d$ dimensions, \acronym{} explanations have size $d^2$, which can be very large. In practice, similarity does not necessarily need to be attributed to every single pair of pixels or input dimensions. A coarse-grained explanation in terms of groups of features jointly representing a super-pixel, a character, or a word, is often sufficient. Let $(\mathcal{I}_1,\mathcal{I}_2,\dots)$ and $(\mathcal{I}'_1,\mathcal{I}'_2,\dots)$ be two partitions of features for the two input examples $\x$ and $\x'$. These partitions form the coarse-grained structure in terms of which we would like to produce an explanation. Coarse-grained relevance scores are then given by:
$$
\textstyle 
R_{\mathcal{I}\mathcal{I}'} = \sum_{i \in \mathcal{I}}\sum_{i' \in \mathcal{I}'} R_{ii'}.
$$
When the original explanation is conservative, it can be verified that the same holds for the coarse-grained explanation ($\sum_{\mathcal{I}\mathcal{I}'}R_{\mathcal{I}\mathcal{I}'} = \sum_{\mathcal{I}\mathcal{I}'} \sum_{ii' \in \mathcal{I}\mathcal{I}'}  R_{ii'} = \sum_{ii'} R_{ii'}$).

\section{Rendering of \acronym{}  Explanations}

\acronym{} explanations of images are composed of $(\# \text{pixels} \times \#\text{pixels})$ scores connecting pairs of pixels in the two input images. Visually rendering these high-dimensional explanations requires to compress them while retaining the relevant information they contain. The rendering procedure we use in this paper is given in Algorithm \ref{alg:rendering}.

\begin{algorithm}
\caption{Rendering of \acronym{} explanations}
\label{alg:rendering}
\begin{spacing}{1.1} 
\begin{algorithmic}
\STATE $R_{\mathcal{I}\mathcal{I'}} \leftarrow \textstyle \sum_{i \in \mathcal{I}}\sum_{i' \in \mathcal{I'}} R_{ii'}$ \hfill {\small (coarse-graining)}
\STATE $R_{\mathcal{I}\mathcal{I'}} \leftarrow R_{\mathcal{I}\mathcal{I'}} / \sqrt[4]{\mathbb{E}[ R_{\mathcal{I}\mathcal{I'}}^4]}$ \hfill {\small (normalization)}
\STATE $ \displaystyle R_{\mathcal{I}\mathcal{I'}} \leftarrow R_{\mathcal{I}\mathcal{I'}} - \text{clip}(R_{\mathcal{I}\mathcal{I'}},[-l,l])$ \hfill {\small (sparsification)}
\STATE $\Delta = h - l$
\STATE $R_{\mathcal{I}\mathcal{I'}} \leftarrow \text{clip}(R_{\mathcal{I}\mathcal{I'}},[-\Delta,\Delta])/ \Delta$ \hfill {\small (thresholding)}
\FORALL{$R_{\mathcal{I}\mathcal{I'}} \neq 0$}
\STATE $\alpha= |R_{\mathcal{I}\mathcal{I'}}|^p$ \hfill {\small (set opacity)}
\IF{$R_{\mathcal{I}\mathcal{I'}} > 0$}
\STATE connect$(\mathcal{I},\mathcal{I}',\text{red},\alpha)$
\ELSE
\STATE connect$(\mathcal{I},\mathcal{I}',\text{blue},\alpha)$
\ENDIF
\ENDFOR
\end{algorithmic}
\end{spacing}
\end{algorithm}

\begin{figure}[h]
\centering
\vskip -2.5mm
\includegraphics[width=1.0\linewidth]{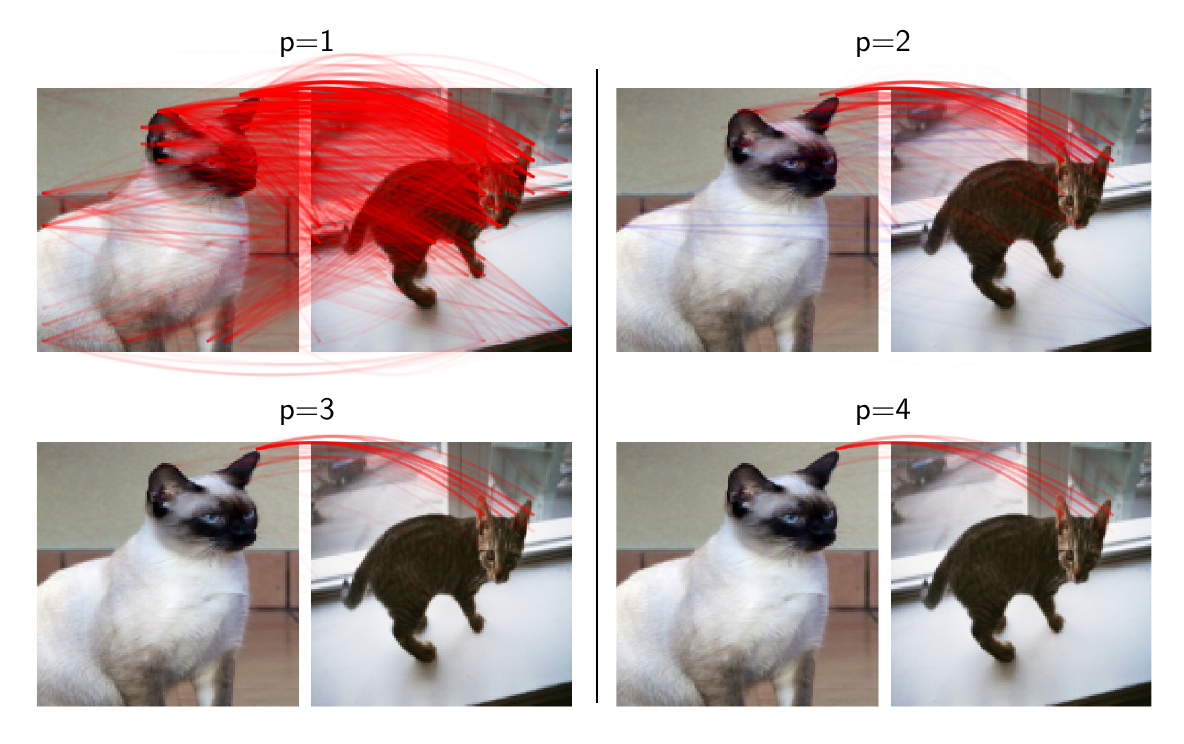}
\vskip -2.5mm
\caption{Effect of the parameter $p$ on the rendering of the explanation. The higher the parameter $p$, the sparser the explanation.}
\label{figure:rendering}
\end{figure}

\noindent The procedure pools relevance scores on super-pixels, normalizes them, shrinks them so that only a limited number of connections need to be plotted, thresholds them so that they fit into a finite color space, and raises them to some power $p$. The parameter $l$ controls the level of sparsification and we tune it mostly for computational reasons. The parameter $h$ forces all scores beyond a certain range to be plotted to the maximum color value. The parameter $p$ lets the explanation focus on all or the highest relevance scores. A large value for $p$ makes it more easily interpretable, however contributions to similarity that are spread to a larger group of input features can become visually imperceptible. Example of heatmaps with different values of $p$ are shown in Fig.\ \ref{figure:rendering}. Parameters retained for each dataset, as well as pooling and input sizes are given in Table \ref{table:parameters}.

\begin{table}[h]
\caption{Parameters used on each dataset for rendering \acronym{} explanations.}
\label{table:parameters}
\small
\centering
\begin{tabular}{lccccc}\toprule
Dataset & input size & pool & $l$ & $h$ & $p$\\\midrule
Pascal VOC 2007 & $128\times 128$ & $8 \times 8$ & 0.25 & 13 & 2 \\
Faces (UFI \& LFW) & $64\times 64$ & $4 \times 4$ & 0.3 & 60 & 1\\
UCF Sport & $128\times 128$ & $8 \times 8$ & 0.25 & 20 &  1\\
Sphaera (illustrations)\!\!\!\!\! & $96\times 96$ & $6 \times 6$ & 0.25 & 15 &  2\\
Sphaera (tables) & $140\times 140$ & $20 \times 20$ & 0.01 & 4 &  2\\
\bottomrule
\end{tabular}
\end{table}

\bibliographystyle{IEEEtran}
\bibliography{similarity}